\documentclass[conference]{IEEEtran}
\usepackage{times}
\usepackage[font=footnotesize]{caption}

\usepackage[numbers]{natbib}
\usepackage{multicol}
\usepackage[bookmarks=true]{hyperref}

\usepackage{color}
\usepackage{graphicx} 
\usepackage{amsmath}
\usepackage{amssymb}
\usepackage{bm}
\usepackage{dsfont}
\usepackage{duckuments}
\usepackage{makecell}
\usepackage{multirow}
\usepackage{subcaption}
\usepackage{placeins}
\usepackage{tikz}
\usepackage{xurl}
\begin{document}

\renewcommand{\tableautorefname}{Tab.}
\renewcommand{\equationautorefname}{Eq.}
\renewcommand{\sectionautorefname}{Sec.}
\renewcommand{\subsectionautorefname}{Sec.}
\renewcommand{\figureautorefname}{Fig.}
\renewcommand{\subsubsectionautorefname}{Sec.}
\title{Map Space Belief Prediction for Manipulation-Enhanced Mapping}

\author{Joao Marcos Correia Marques$^{1*}$ \quad Nils Dengler$^{2,3,4*}$\quad Tobias Zaenker$^{2,4}$\quad Jesper Mücke$^{2}$ \\ Shenlong Wang$^1$ \quad Maren Bennewitz$^{2,3,4}$ \quad Kris Hauser$^1$
\\ \small $\ast$ These authors contributed equally to this work \\ 1. University of Illinois at Urbana-Champaign, IL, USA \quad 2. Humanoid Robots Lab, University of Bonn, Germany \\ 3. The Lamarr Institute, Bonn, Germany \quad 4. The Center for Robotics, University of Bonn, Germany
}

\makeatletter
\g@addto@macro\@maketitle{
\setcounter{figure}{0}
\centering
\begin{tikzpicture}
  \node (img) {\includegraphics[width=\linewidth]{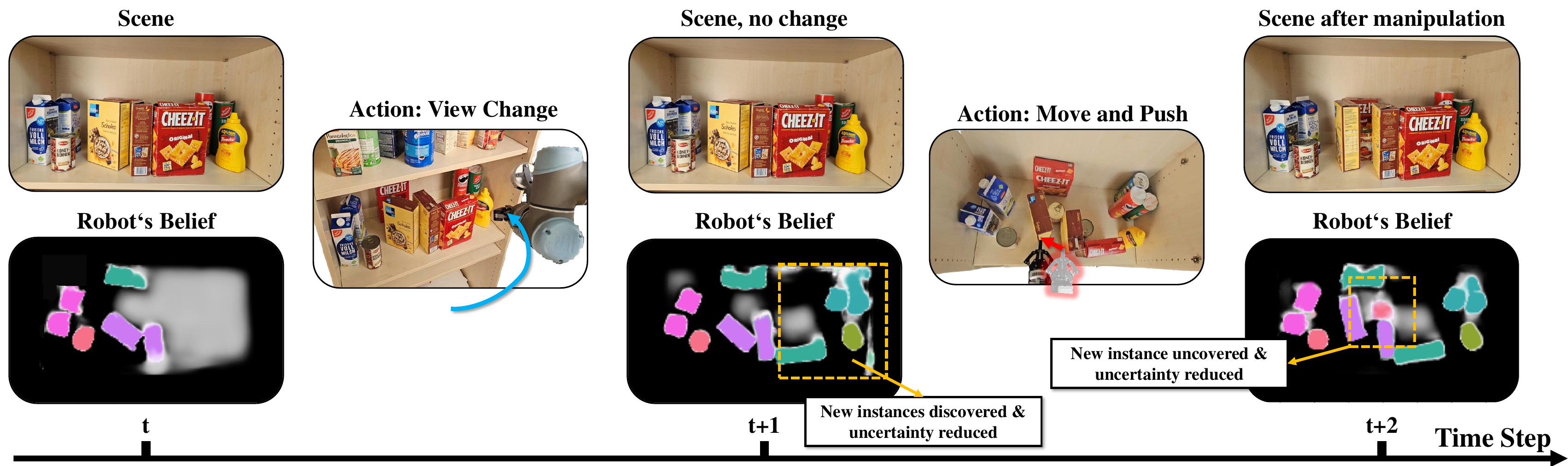}};
\end{tikzpicture}
	\captionof{figure}{Example scenario with occlusions in a confined shelf environment. 
Given a current partial map of the environment~(belief~$t$), our planner decides whether gathering another observation or manipulating the scene would be best to reduce map uncertainty. In this example, first a observation action would increase environmental knowledge, followed by a push to unveil the hidden can behind the two boxes at time $t+2$. The predicted belief map is visualized as a top-down projection of the shelf, ignoring the occluding top shelf board.}
 \label{fig:coverfig}
	\vspace{-2ex}
  
}
\makeatother

\maketitle

\begin{abstract}
Searching for objects in cluttered environments requires selecting efficient viewpoints and manipulation actions to remove occlusions and reduce uncertainty in object locations, shapes, and categories.
In this work, we address the problem of manipulation-enhanced semantic mapping, where a robot has to efficiently identify all objects in a cluttered shelf.
Although Partially Observable Markov Decision Processes~(POMDPs) are standard for decision-making under uncertainty, representing unstructured interactive worlds remains challenging in this formalism.
To tackle this, we define a POMDP whose belief is summarized by a metric-semantic grid map and propose a novel framework that uses neural networks to perform map-space belief updates to reason efficiently and simultaneously about object geometries, locations, categories, occlusions, and manipulation physics. 
Further, to enable accurate information gain analysis, the learned belief updates should maintain calibrated estimates of uncertainty. 
Therefore, we propose Calibrated Neural-Accelerated Belief Updates (CNABUs) to learn a belief propagation model that generalizes to novel scenarios and provides confidence-calibrated predictions for unknown areas. 
Our experiments show that our novel POMDP planner improves map completeness and accuracy over existing methods in challenging simulations and successfully transfers to real-world cluttered shelves in zero-shot fashion.

\end{abstract}

\IEEEpeerreviewmaketitle

\section{Introduction}
Active sensing has long been studied in robotics for tasks such as exploring an unknown environment~\cite{bircher2016three}, complete 3D object model acquisition~\cite{krainin2011autonomous}, and searching for an unobserved target object~\cite{li2016iros,georgakis2022learning, Zhai_2023_ICCV}. 
To build complete maps as efficiently as possible, Next Best View (NBV) planning~\cite{zeng2020view} is often employed to reduce the uncertainty about the map as quickly as possible. 
Although NBV planning offers an approach for static scenes in which the robot simply moves the camera passively through free space, there are many applications, such as household and warehouse robotics, in which robots may need to manipulate the environment in order to gain better viewpoints~\cite{dengler2023viewpoint, pitcher2024iros}. 
We refer to this problem as {\em Manipulation-Enhanced Mapping} (MEM).
MEM offers two significant new challenges beyond standard NBV problems. 
First, in order to decide when and where to manipulate objects, the robot should reason about how object movement may affect previously occluded regions. 
Second, it must anticipate the impact of manipulations on observed objects and possibly partially-observed or unobserved objects. 
For example, pushing boxes in a grocery shelf backward will move all objects simultaneously until the furthest, occluded one, hits a wall. 

MEM is related to the {\em mechanical search} problem \cite{danielczuk2019mechanical} in which the robot manipulates clutter to reveal a target object. 
Prior approaches in mechanical search tend to rely on restrictive assumptions, such as a static viewpoint, which ignores a robot's ability to look around obstacles~\cite{huang2021mechanical,sharma2023openworld}. 
Other studies assume full observability of object dynamics and poses \cite{saxena2024improved} or are limited to a fixed set of predefined objects \cite{xiao2019online}. 
These assumptions are too limiting for complex cluttered scenes like shelves. 
The most closely-related work to ours is \citet{dengler2023viewpoint}, who address these limitations by training a reinforcement learning policy for viewpoint planning and learn a push scoring network from human annotations to derive manipulation actions, switching between manipulation and manipulation when the information gain from obtaining novel images of the environment saturates.
However, their approach to switching between action modes is inefficient, since waiting for information gain saturation to perform a push results in the agent sampling the environment several times to reveal details that could have been revealed easily through manipulation. 
Furthermore, the proposed method does not update its environmental map after a push, leaving the viewpoint planner, conditioned solely on this outdated map, less capable of exploiting the newly revealed space.

This paper formulates the MEM problem as a Partially Observable Markov Decision Process (POMDP) in the belief space of semantic maps. By maintaining map-space beliefs, our approach is applicable to unstructured cluttered environments with an arbitrary numbers of objects.
The POMDP computes the next best viewpoint or manipulation action that maximizes the agent's expected information gain over a short horizon (\autoref{fig:coverfig}) in a receding-horizon fashion.
Our approach leverages neural network methods for map-space belief propagation, which have been shown in the object goal navigation literature to drastically improve map completion rates and offer better guidance for object search \cite{georgakis2022learning, Zhai_2023_ICCV}. 
The key challenge in belief propagation with manipulation actions is that they often reduce certainty when the object's dynamics are unknown or the robot interacts with unobserved objects.
To address this challenge, we introduce the Calibrated Neural-Accelerated Belief Update (CNABU) technique to learn belief propagation models for both observation (obtaining new images from the environment) and manipulation actions. 
Confidence calibration is especially important for belief propagation because overconfidence in either object dynamics or map prediction can result in ineffective exploration and/or early termination. 
We employ evidential deep learning to obtain better off-the-shelf model calibration~\cite{NEURIPS2018_a981f2b7}. 

Our experiments in simulation environments demonstrate that our proposed MEM planner outperforms prior work \cite{dengler2023viewpoint} and CNABU-enhanced baselines in terms of metric-semantic accuracy.
Furthermore, we perform hardware experiments with a UR5 robot equipped with a gripper and an in-hand camera, demonstrating zero-shot transfer of the learned models, and showing the efficacy of our method in mapping of cluttered shelves. 
An implementation of our method can be found on
Github\footnote{\url{https://github.com/NilsDengler/manipulation_enhanced_map_prediction}}.

\section{Related Work}
\subsection{Next Best Viewpoint Planning}
NBV planning is a well-researched approach in the area of active vision that has been applied to both object reconstruction and large-scale scene mapping.
Generally, NBV consists of two steps: First sampling view candidates, then evaluating which candidate is the best.
For object reconstruction tasks like \cite{Hu24icra}, views are usually sampled from a fixed set around the object.
For large-scale scenes, sampling is more challenging. \citet{monica2018contour} sample at the contour of the explored scene.
Other approaches sample at either predefined or dynamically detected regions of interest.
For the evaluation, most approaches compute an estimated information gain to determine the utility of a view.
The information gain is often based on the expected entropy reduction, e.g. by counting unknown voxels in the field of view.
Other approaches like Hepp et al. \cite{hepp2018learn} rely on a learned utility to predict the best view.
In this work, we build upon existing concepts of NBV planing, but enhance them by incorporating manipulation actions to interactively shape and explore the environment, allowing the robot to gather richer information and adapt its strategy based on both observation and interaction.

\subsection{Mechanical Search in Shelves and Piles}
Mechanical search algorithms \cite{danielczuk2019mechanical, huang2021mechanical,sharma2023openworld} locate and extract one or multiple target objects from a given scene, while dealing with confined spaces, occlusion and object occurence correlations. 
The task consists of multiple steps, i.e., visual reasoning, motion and action planning as well as their precise execution. 
For visual reasoning, current research demonstrates that the scene can be effectively explored by interacting with objects~\cite{bohg2017interactive, dengler2023viewpoint,li2016iros, pitcher2024iros} to actively reduce or overcome occlusions, but most works consider a fixed viewpoint \cite{huang2021mechanical,danielczuk2019mechanical,sharma2023openworld}. 

\citet{kim2023corl} propose a method for locating and retrieving a target object using both pushing and pick-and-place actions.
However, their approach relies on a fixed camera, lacks a long-term map, and rebuilds environmental knowledge from scratch with each observation. Therefore, the approach can lead to unnecessary manipulation actions, as the target may already be visible from other viewpoints.
In the context of planning for Manipulation Among Movable Objects (MAMO) \cite{saxena2023planning, stilman2007manipulation}, \citet{saxena2024improved} introduced a method for object retrieval in cluttered, confined spaces. 
Despite achieving strong retrieval performance, their approach depends entirely on prior knowledge of object shapes and dynamics.
\citet{pomdp_compositional_uncertainty} propose a POMDP formulation for manipulating objects under uncertain segmentation using a particle belief representation, but limit their analysis to fixed viewpoints and prehensile manipulation to enable efficient belief propagation.
\citet{visibility-aware_manipulation} propose a planner based on two-level hierarchical search to enable visibility-aware navigation with movable objects. 
Their algorithm, Look and Manipulate Backchaining (LAMB), however, relies on the assumptions of deterministic action outcomes and on extremely simplified environment dynamics (grasps always succeed, pushed objects always slide along axis without rotation), having limited applicability in the real world.

In this work, we do not focus on retrieving individual objects, but on mapping and identifying all objects within an environment. 
With our long-term occupancy and semantic map representation, retrieval plans can be generated without relying on perfect model knowledge or single-shot scene understanding, while our learned belief updates enable modeling of more complex manipulation behaviors.

\subsection{Learned World Dynamics Models}
Many model-based reinforcement learning algorithms learn environment models from episodic environmental interaction, often in latent spaces for improved evaluation speed~\cite{Hafner2020Dream}. 
These models, however, do not result in a human-interpretable representation in contrast to our learned map-space dynamics. 
Another recent trend leverages the popularity of conditional video diffusion models to develop interactive ``pixel-space'' simulators~\cite{yang2024learning,valevski2024diffusionmodelsrealtimegame}. 
These models focus on short-term visual fidelity.  We posit that explicit 3D maps provide long-term temporal consistency and calibrated uncertainty measures that are needed for robotic tasks that involve information gathering.

\section{Problem Definition}
We address the MEM problem as follows:
Consider an environment with a set of movable objects of varying sizes and orientations, where some objects may be occluded and not directly observable from any viewpoint.
The arrangement of these objects, along with the fixed support geometry (e.g., a shelf or table), constitute the workspace 
configuration space~$C^w$ \cite{10.1007/978-3-030-66723-8_5}. 
The environment's initial configuration is unobservable and denoted $c_0 \in C^w$.

The robot's objective is to create an accurate representation of the workspace configuration $c_t$ after the execution of a sequence $\zeta$ of actions $[a_0,\ldots, a_n]$.
These actions include two types: {\bf observation actions} where the robot moves its camera to a specific viewpoint $v_t\in \mathds{V}$ to capture an RGB-D image, and {\bf manipulation actions}, where the robot interacts with the scene~(e.g., via pushing or grasping). We address the eye-in-hand RGB-D camera setting in which the robot does not receive informative observations during manipulation and must instead move to a retracted viewpoint to receive valid depth data due to minimum depth restrictions. Moreover, we do not integrate views during movement between locations, since such images are subject to motion blur.

To formalize the problem, let the robot's internal representation of the environment, a belief over metric-semantic maps explaining object classes over the workspace, be denoted as $\Phi_t$.  We assume a closed set of $N_{\mathrm{classes}}$ semantic classes.  Let the most-likely map according to a belief $\Phi_t$ be denoted $\phi_t$.
When the robot executes a manipulation action \textbf{$a_t$}, it causes a transition ${c_t \mapsto c_{t+1} \in C^w}$ according to the environment's dynamics $c_{t+1} = \mathrm{Dyn}(c_{t}
,a_t)$. 
Additionally, whenever the robot takes any action $a_t$ and gets an observation~\textbf{$o_t$}, drawn according to the observation function $Z(o_t|a_t,c_{t+1})$, its internal representation is updated through its belief update function $\Phi_{t+1} = \mathrm{BelUpdate}(\Phi_{t},a_t,o_t)$. 

Finally, similar to \citet{7989112}, we define MEM as an optimal budgeted mapping problem. 
The robot is given a maximum action budget \textit{T}, and an initial environment configuration $c_0$, which is \textit{a priori} unknown. The task is to output the most informative sequence of actions $\zeta$ such that the robot's predicted map $\phi_T$, at the last step of the budget, maximizes its mean Intersection over Union~(mIoU) to the ground truth map $\phi^{\mathrm{GT}}_T$ which represents $c_T$. 
So we have:
\begin{equation}
\label{eq:obm_definition}
\begin{aligned}
  \zeta^* \quad &= \underset{{\zeta = [a_0,\cdots,a_{T-1}]} }{\arg\max}\text{mIoU}\left(\phi_T,\phi^{\mathrm{GT}}_T\right) \\
  \textrm{s.t.} \quad & \phi^{\mathrm{GT}}_T = \textrm{ToMap}(c_T) \\
  \quad & c_{t+1} = \textrm{Dyn}(c_{t},a_t) \forall t\\
  \quad & \Phi_{t+1} = \textrm{BelUpdate}(\Phi_{t}|a_t,o_t), o_t \sim Z(o_t|a_t,c_{t+1}), \forall t
\end{aligned}
\end{equation}
where $\mathrm{BelUpdate}(\cdot)$ represents the robot's belief update, $Z(\cdot)$ is the observation function, and $\mathrm{ToMap}(\cdot)$ yields the metric-semantic map that corresponds to a known configuration of the environment. In deployment, the robot cannot accurately predict $\phi^{\mathrm{GT}}_T$, as it does not have access to the initial configuration nor the dynamics of the environment. It may not even know the number of objects, their shapes, or semantic labels.

\section{Method}
\vspace{-5px}
\subsection{Overview}
We model the MEM problem as a Partially Observable Markov Decision Process (POMDP) in metric-semantic map-space $\Omega$. To solve this POMDP, the agent should perform a belief update about the state of the map after both manipulation and observation actions.
For manipulation actions, the belief update propagates through a map transition function $T(\phi_{t+1}|\phi_t,a_t)$ that is \textit{a priori} unknown.  For observation actions, the belief update should integrate the observation while reasoning over hidden object shapes and arrangements to reduce uncertainty.
\begin{figure*}[!ht]
	\centering
\includegraphics[width=0.99\linewidth, trim={6cm 5cm 6cm 4cm},clip]{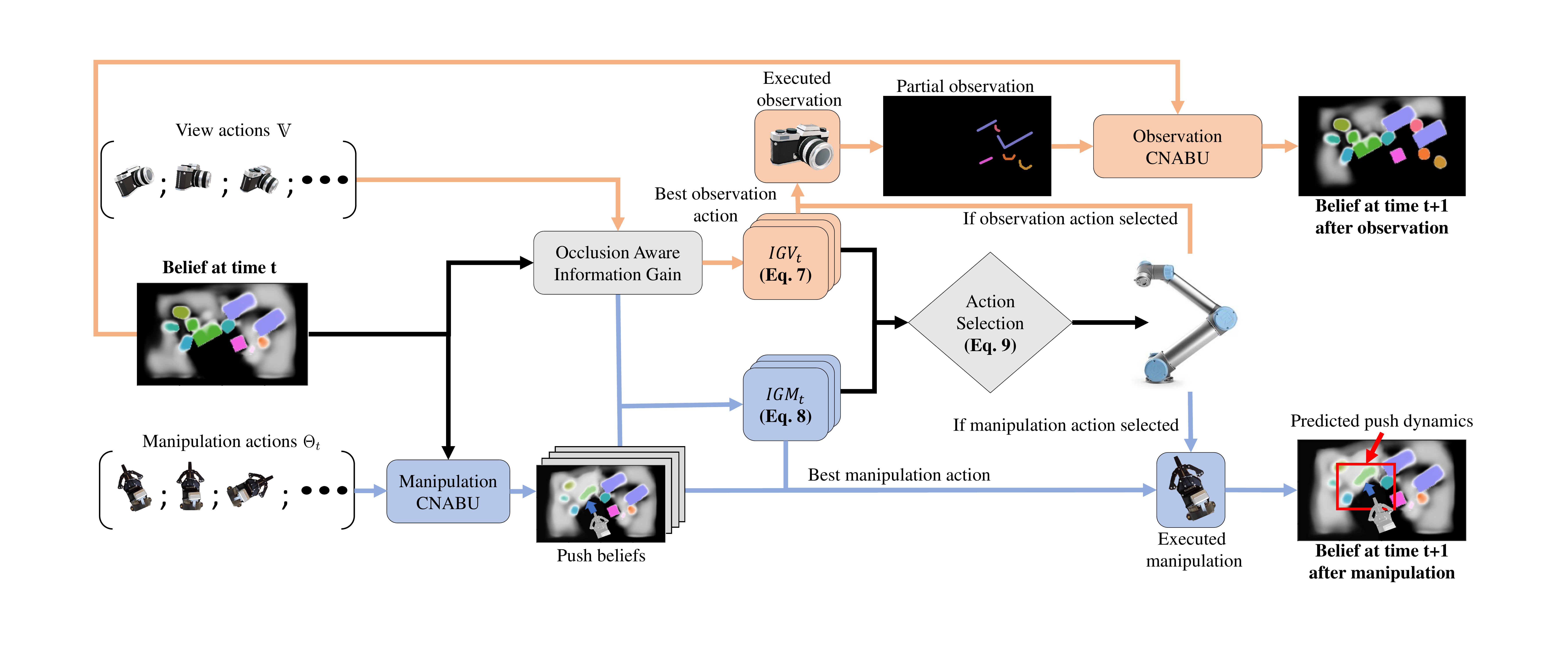}
	\caption{From a prior map belief, our pipeline predicts a map belief resulting from a set of candidate pushes. 
  It then weighs the information gain from taking two consecutive independent views given the current belief (orange arrows) or taking a single observation given any of the predicted beliefs after pushing (blue arrows), selecting the path of highest cumulative information gain and taking its respective first action -- either taking the next best view or executing the best push. $IGV_t$ represents the best information gain obtainable from taking two distinct observation actions, whereas $IGM_t$ is the best information gain obtainable through a manipulation action followed by an observation action. 
	} 
	\label{fig:framework_illustration}
\vspace{-10px}
\end{figure*}

However, the high dimensionality of the space of all possible maps makes traditional belief updates computationally infeasible \cite{KAELBLING199899}. The approximation that  map cells are independent, as commonly done in occupancy grid mapping \cite{thrun2005probabilistic}, leads to compact belief representations but reduces precision in the belief update.  Our CNABU method captures prior knowledge about the world, such as object contiguity and usual object sizes and arrangements, in its update.
We also leverage uncertainty-aware deep learning models to predict factorized belief updates that are better aligned with plausible map configurations and yield actionable quantification of uncertainty. 

These models are trained using simulated ground truth to approximate occlusion reasoning and interaction dynamics, \textit{i.e.}, $\mathrm{Dyn}$. Object sizes, classes, occlusion levels, and manipulation effects should be roughly representative of real scenes, but our method is tolerant to differences in configuration, number, class distribution, and moderate shape changes. 

Finally, our POMDP solver approximates the MEM objective with the Volumetric Information Gain (VIG) metric \cite{Delmerico2018AReconstruction}, since directly modeling the mIoU metric is not feasible, as it depends on the unobservable workspace configuration.
It uses a 2-step greedy approach that obtains good performance by exploiting near-submodularity of the VIG function. Moreover, the 2-step approach obviates the need to sample from the observation distribution $Z$.

\subsection{Neural Map Belief Dynamics}
\label{sec:map_completion_network}
Following grid mapping literature \cite{thrun2005probabilistic}, we represent a belief $\Phi_t$ over the semantic-metric map using a Bernoulli distribution for a cell's occupancy and a categorical distribution for a cell's semantic class. Each cell is assumed independent. We represent the occupancy map as a 3D voxel belief $\Phi_t^O \in \mathbb{R}^{H \times W \times D}$ and the semantic map as a 2D birds-eye belief $\Phi_t^S \in \mathbb{R}^{H \times W \times N_{classes}}$. The semantic map is 2D for simplicity because objects are roughly prismatic and stacking is not allowed in our problem, while the occupancy is 3D because object heights affect visibility determination.
We consider observations $o_t\in O$ consisting of an RGB-D image with added semantic labels.

To perform efficient belief updates, we introduce a Calibrated-Neural Accelerated Belief Update (\textbf{CNABU}) technique that uses separate neural networks to represent the posteriors of the viewpoint and manipulation actions in the factored representation.  The first, called {\em observation CNABU}, computes a map belief update after a observation action $\Phi_{t+1} \gets \sigma_o(\Phi_t,o_t,a_t)$.  The second, called {\em manipulation CNABU}, computes a map belief update after a manipulation action $\Phi_{t+1} \gets \sigma_m(\Phi_t,a_t)$, where we drop the observation because no new observation is generated.

Let $\Phi_t(\phi) = P(\phi|\Phi_t)$ denote the probability density of a map $\phi$ under belief $\Phi_t$. 
For any action $a_t$ and observation $o_t$, the standard POMDP belief update equation~\cite{KAELBLING199899} gives the posterior belief as:
\begin{equation}
  \label{eq:traditional belief update equation}
  \Phi_{t+1}(\phi) = \frac{1}{\eta} Z(o_{t}|a_{t},\phi)\sum_{\phi' \in \Omega} T(\phi|\phi',a_t)\Phi_{t}(\phi'),
\end{equation}
where $\phi'$ ranges over all possible maps, with $\eta$ being a normalizing constant. If we wished to project the belief state into a marginalized occupancy probability of a cell, we would need to compute:
\begin{equation}
  \label{eq:independent updated factorization}
  \Phi_{t+1}^O[i,j,k] = \mathds{E}_{\phi\thicksim \Phi_{t+1}}[\phi^O[i,j,k] = 1],
\end{equation}
where  $\mathds{E}$ is the expected value.  A similar equation would hold for semantic updates.  Regardless, the space of possible maps is far too large \cite{thrun2005probabilistic} and there is no current method for assessing map densities $\Phi_t(\phi)$ that accurately accounts for inter-cell correlations (e.g., object shapes, arrangements, and visibility). 

Instead, we train CNABUs through simulation data.
Since any scene that could produce a belief $\Phi_t$ via its observations is inherently a sample of the distribution induced by this belief, the training process leverages a neural network's averaging tendency to create an implicit Monte Carlo estimate of \autoref{eq:independent updated factorization}. We postpone the network and training details to Section.~\ref{sec:NetworkTraining} and proceed to describe their use in MEM.

\begin{figure*}[!ht]
	\centering
	\includegraphics[width=0.99\textwidth, trim={0cm 0cm 0cm 0cm},clip]{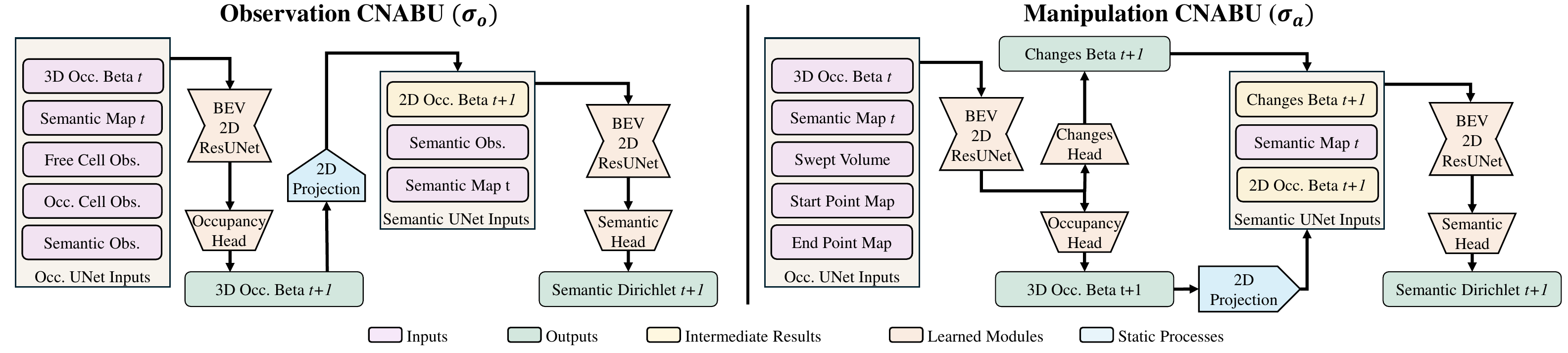}
	\caption{Architecture overview of our observation and manipulation CNABU networks. The observation prediction network uses the occupancy posterior beta and semantic posterior Dirichlet for loss computation, while the manipulation prediction network additionally takes the 2D map of occupancy changes after the push for loss calculations.
	} 
	\label{fig:net_architecture}
\vspace{-10px}
\end{figure*}

\subsection{Solving the POMDP}
\label{sec:solving_the_pomdp}
We propose to solve the map-space POMDP by using a $k$-step receding horizon greedy planner, as shown in \autoref{fig:framework_illustration}, which uses 
Volumetric Information Gain (VIG) \cite{Delmerico2018AReconstruction} as an approximation of the true reward. VIG correlates to information gain and is a submodular optimization objective in static scenes \cite{krause2008near}, having been used to efficiently solve NBV planning and sensor placement problems.
Due to VIG's submodularity, a greedy policy for solving this problem leads to bounded suboptimality, justifying the greedy receding-horizon strategy.
While VIG's submodularity does not hold in general for dynamic scenes, we assume that manipulation actions do not increase entropy at a rate that justifies the significant expense of long lookaheads. However, because a manipulation action does not produce any observation (and hence no immediate information gain) at least one further observation must be considered in scoring. 
Hence, the minimum viable search horizon at $k = 2$, which yields a balanced tradeoff between computational efficiency and action quality.

To perform an observation action, the robot chooses from $v^i \in \mathds{V}$ possible views in a fixed array of camera positions $\mathds{V}$ to which the robot can move. 
Furthermore, let $\theta_t \in \Theta_t$ be a sampled manipulation action from a set of feasible manipulation actions. 
In our two-step greedy search, we only consider two possible kinds of action sequences: taking two observation actions $(v_t,v_{t+1})$ or performing a manipulation action followed by an observation $(\theta_t,v_{t+1})$. 
This is because $(\theta_t,\theta_{t+1})$ would result in no observation and therefore no information gain and the information gain of $(v_t,\theta_{t+1})$ is strictly smaller than the VIG of any $(v_t,v_{t+1})$, $v_{t+1}\neq v_t$.

Letting $IG(v_i,\cdots,v_{i+n-1}|\Phi^O)$ denote the Volumetric Occlusion-aware Information Gain \cite{Delmerico2018AReconstruction} for voxels intersected by the rays from $n$ views $\left(v_i\cdots v_{i+n-1}\right)$ given belief $\Phi^O$
, the two most informative consecutive views $(v_t^*,v_{t+1}^*),$ are:
\begin{equation}
  (v_t^*,v_{t+1}^*) = \underset{v_t,v_{t+1}\in \mathds{V}}{\arg\max}IG(v_t,v_{t+1}|\Phi_t^O)
\end{equation}

For manipulation, $\tilde{\Phi}_{t+1}^{\theta_t} = \sigma_m(\Phi_t,\theta_t)$ denotes the predicted belief from the manipulation CNABU when given action \mbox{$\theta_t \in \Theta_t$} as input.
We can define the most informative 1-step push, $\theta_{t}^*$ and its associated most informative view $v_{\theta_t}^*$, as:
 \begin{equation}  
 \left(\theta_{t}^*,v_{\theta_t}^*\right) = \underset{\theta_t\in \Theta_t,v_{t+1}\in V}{\arg\max} IG(v_{t+1}|\tilde{\Phi}_{t+1}^{\theta_t})
 \end{equation}
Finally, let $H(\cdot)$ represent the entropy of a given semantic map, with the entropy change defined as \mbox{$\Delta H(\Phi_t, \Phi_{t+1}) = H(\Phi_t) - H(\Phi_{t+1})$}. 
The term \mbox{$IGV_t = IG(v^*_t, v^*_{t+1} | \Phi_t^O)$} denotes the best information gain obtained from two viewpoints, also called \textbf{ViewPoint Planning} (VPP), while \mbox{$IGM_t = IG(v^*_{\theta_t} | \tilde{\Phi}_{t+1}^{\theta_t^*})$} represents the best information gain from a push followed by a viewpoint, which we will call \textbf{push selection}.
Lastly, \mbox{$Reg_t = \Delta H(\Phi_t, \tilde{\Phi}_{t+1}^{\theta^*_t})$} captures the entropy difference between the current semantic map and the map after the best push. 

Our policy decides the action $a_t$ to take according to:


\begin{equation}
a_t = \begin{cases}
v_t^* &\text{if } IGV_t > IGM_t + \gamma Reg_t\\
\theta_t^* &\text{ otherwise } 
\end{cases}
\end{equation}

Where $\gamma$ is a discount factor to account for different magnitudes between camera array VIGs and whole-map semantic entropy values. 
$Reg_t$ serves as a regularization on the aggressiveness of the selected pushes, as more radical environment manipulations could potentially reveal more of the environment, but would cause unnecessary disturbances to the scene and introduce a lot more uncertainty to the post-push representations.
If $a_t = \theta^*_t$, no observation is taken and $\Phi_{t+1} = \tilde{\Phi}_{\theta_t^*, t+1}$. If $a_t$ is an observation action, we get the observation $o_{t}$ and use $\sigma_o$ to obtain the new belief $\Phi_{t+1} = \sigma_o(\Phi_t,a_t,o_t)$. 
This search is then repeated in a loop until the maximum number of actions has been performed, or a threshold for full map completion has been reached, which we set to 95\% of the semantic voxels with greater than 85\% certainty in the belief $\Phi$. 
When this threshold is reached, the planner no longer pushes, but still collects novel views. 

\subsection{Push Sampling}
\label{sec:push_sampling}
We consider pushing as our manipulation action. To compute valid push candidates using $\Phi_t^O$, we first compute the high-confidence frontier points from the shelf entrance and sample k of them uniformly at random as start points for the pushes. We test the start points of the k sampled pushes for collisions against other high confidence voxels in $\Phi^O_t$. 
After randomly sampling these $k$ unique frontier points, we determine for each point whether this starting position of a push will lead to a feasible and valid motion plan. 
For each valid start point, we sample a likely occupied point in $\Phi^O_t$ near it to obtain the push direction and sample a push distance uniformly at random between 50 and 150\,mm. 
We then obtain a valid motion plan using a sampling based motion planner \cite{Gildardo2003AChecking,hauser13klampt} and parametrize this plan with $\theta$.

\subsection{Evidential Posterior Learning}

Although we could learn to predict beliefs as functions of $\Phi$, we introduce an enhanced representation that uses evidential posterior networks~\cite{ulmer2023prior}, as evidential learning is known improve the calibration of uncertain predictions. An {\em evidential map belief} $\bm{\lambda}$ consists of belief prior parameters for each cell of the map belief $\Phi$. Specifically,  we store $\bm{\lambda^O} \in \mathds{R}^{H \times W \times D \times 2}$, a 3D grid of Beta distribution parameters for each voxel in the map, and $\bm{\lambda^S} \in \mathds{R}^{H \times W \times N_{\mathrm{classes}}}$, a grid of Dirichlet distribution parameters for each cell in the 2D map.
Let $\mathrm{Beta}(\cdot)$ and $\mathrm{Dir}(\cdot)$ denote the Dirichlet and Beta distributions, respectively.
Therefore, the occupancy and semantic map beliefs are related to evidential parameters via $\Phi^O[\cdot] = \mathds{E}[\mathrm{Beta}(\bm{\lambda^O}[\cdot])]$ and $\Phi^S[\cdot] = \mathds{E}[\mathrm{Dir}(\bm{\lambda^S}[\cdot])]$ . We assume that the initial states $\bm{\lambda_0^O}$ and $\bm{\lambda_0^S}$ are uninformed and set to $\mathds{1}$, a unit tensor with appropriate dimensions.  Note: evidential parameters are maintained for each map belief $\Phi$ in our algorithm and CNABUs operate by propagating the evidential parameters ($\bm{\lambda}_{t+1} \gets \sigma_o(\bm{\lambda_{t}},o_t,a_{t})$ and $\bm{\lambda}_{t+1} \gets \sigma_m(\bm{\lambda_{t}},a_{t})$) followed by an update to the standard belief parameters.

\subsection{Dataset Generation}
\label{sec:viewpoint_cnabu_dataset_generation}

To train the CNABU models $\sigma_o$ and $\sigma_m$, we collect datasets on maps, viewpoints, and sampled pushes in simulation. A total of 14 different object categories from the YCB dataset \cite{calli2015benchmarking} are used and sampled in a shelf board of size $(0.8 \times 0.4 \times 0.4)m$. 
We sample object configurations on the shelf following a stochastic method that considers class dependencies and efficient free space coverage for placing objects. 
This method allows for the sampling of varied object configurations, numbers and classes, and is described in more detail in Appendix \ref{sec:dataset_generation}.

To train the viewpoint belief prediction model $\sigma_o$, we assemble a dataset $\mathds{D} = \{d_1,d_2,\ldots\}$ where each scene $d_i$ has the form $(\phi^{\mathrm{GT}}, o_1, \cdots, o_n)$. 
Here, $\phi^{\mathrm{GT}}$ represents the ground truth 3D metric-semantic voxel map of a shelf environment with randomly placed objects, and $o_1, \cdots, o_n$ are the depth and semantically segmented images captured from $\mathds{V}$, the set of discrete viewpoints in the environment. The ground truth semantic labels are used in the rendered images.
 
To train the manipulation belief prediction model $\sigma_m$, the simulated robot executes a randomly sampled action in synthesized scenes. 
This produces a dataset $\mathds{D}^a = \{d^a_1,d^a_2,\ldots\}$ with each sequence $d^a_i = (\phi^{\mathrm{GT}}_{\mathrm{pre}}, \phi^{\mathrm{GT}}_{\mathrm{post}}, o_1, o_2, \cdots,o_n,a)$, where $\phi^{\mathrm{GT}}_{\mathrm{pre}},\phi^{\mathrm{GT}}_{\mathrm{post}}$ are the ground truth maps before and after manipulation, respectively, $a$ is the executed action and $o_1,\cdots,o_n$ are observations from $\mathds{V}$ as before, taken before the manipulation is executed.

\subsection{Training CNABU Networks}
\label{sec:NetworkTraining}

We now outline the procedure for training the two CNABUs, whose network architectures are shown in \autoref{fig:net_architecture}, with further details given in Appendix~\ref{sec:cnabu_implementation_details}. 

We train $\sigma_o$ from the dataset $\mathds{D}$ by sampling, without replacement, a sequence of $l$ posed depth and semantic images, $d' = (o'_0,\cdots, o'_{l-1})$, from every scene $d_i$ at every epoch. 
This sampling diversifies the beliefs encountered during training by varying the emulated observation sequences for each scene.

The loss over the sequence $d'$ is computed as follows.
We recursively predict the evidential beliefs $\bm{\lambda_t} =\sigma_o(\bm{\lambda_{t-1}},o'_{t-1})$ up to time $l$. Also, let $y^i\in\{0,1\}^{\mathrm{dim}}$ indicate a one-hot encoded tensor of the ground truth value for a given voxel i according to $\phi^{\mathrm{GT}}$, where $\mathrm{dim}$ is either $N_{\mathrm{classes}}$ for $\mathbf{\lambda^S_t}$ and 2 for $\mathbf{\lambda^O_t}$. 
For each voxel i in $\phi_t$, define, for its predicted distribution parameter $\lambda_t^i \in \bm{\lambda}_t$, $\tilde{\lambda}_t^i=y^i + (1-y^i)\odot\lambda_t^i$, where $\odot$ denotes element-wise multiplication.
We employ the evidential uncertainty-aware cross-entropy from \citet{NEURIPS2018_a981f2b7} as the loss, which, for each time step, is given by:

\begin{equation}
  \label{eq:uncertainty_aware_cross_entropy_loss}
  \resizebox{0.91\linewidth}{!}{%
  $L^{\mathrm{type}}_t(\mathbf{\lambda_t^{\mathrm{type}}},\phi^{\mathrm{GT}})= \sum\limits_{\lambda_t^i\in\mathbf{\lambda_t^{\mathrm{type}}}}\mathcal{L}(\lambda^i_t,y^i) + \varepsilon \mathrm{KL}\left(Dir(\tilde{\lambda}_t^i)||Dir(\mathds{1})\right)$

  }
\end{equation}
, where $ \mathcal{L}(\lambda_t^i,y^i) = \sum_{j = 0}^{\mathrm{dim}}y^{ij}\left(\text{log}(S^i_t)-\text{log}(\lambda^{ij}_t)\right)$, $S^i_t = \sum_{j=0}^{\mathrm{dim}}\lambda_t^{ij}$, $\mathrm{KL}(a||b)$ is the Kullback-Liebler divergence between distributions $a$ and $b$, $\mathds{1}$ is a vector of all ones, $type$ is $o$ if it is an occupancy loss (dim = 2) and $s$ if it is a semantic loss (dim = $N_{\mathrm{classes}})$, and $\varepsilon$ is an annealing parameter, set according to \citet{NEURIPS2018_a981f2b7}.  The total loss for the sample $d'$ is the sum of the semantic and occupancy losses $L^{o}_{t}+L_{t}^s$ over the $l$ observations.

The manipulation CNABU is defined similarly to the observation CNABU, except that it has an auxiliary output, which predicts a Beta distribution over a voxel grid modeling the probability of a given voxel being changed in $\Phi^{\mathrm{GT}}$ after the manipulation is executed, which we call $\bm{\lambda^{\mathrm{diff}}}$. Therefore, we have $\bm{\lambda^S_{t+1}}\bm{\lambda^O_{t+1}},\bm{\lambda^{\mathrm{diff}}_{t+1}} = \sigma_m(\bm{\lambda}_t,a_t)$

Training epochs iterate over sequences in $\mathds{D}^a$.  For each sequence, we sample a subsequence of $l\in[1,10]$ images without replacement as above. We then recursively obtain the beliefs from the observation CNABU, ${\bm{\lambda}_{t+1} = \sigma_o(\bm{\lambda}_{t},o'_t)}$ for $t=0,\ldots,l-1$. Next, we predict the post-manipulation belief $\bm{\lambda^S_{t+1}}\bm{\lambda^O_{t+1}},\bm{\lambda^{\mathrm{diff}}_{t+1}} =\sigma_m(\bm{\lambda}_l,a)$ and use $\phi^{\mathrm{GT}}_{\mathrm{post}}$ as the training target. The random sampling of a different number of observations prior to pushing ensures the CNABU sees different belief stages during training. We derive the ground truth for $\bm{\lambda^{\mathrm{diff}}}$, $\phi_{\mathrm{change}}^{\mathrm{GT}}$, from the difference between $\phi_{\mathrm{pre}}^{\mathrm{GT}}$ and $\phi_{\mathrm{post}}^{\mathrm{GT}}$.

Finally, we add a fourth loss component, which we call consistency loss, $L^{\mathrm{con}}$ which is the Mean Squared Error between $\bm{\lambda_{l+1}}$ and $\bm{\lambda_{l}}$. This loss serves as a regularization to encourage the alpha parameters of the distributions in the unchanged areas of the map to have a similar magnitude.
 As before, the network heads are trained using the loss in \autoref{eq:uncertainty_aware_cross_entropy_loss}~\cite{NEURIPS2018_a981f2b7}. The total loss for the manipulation sequence is given by $L = L_{l+1}^O+L_{l+1}^S+L_{l+1}^{\mathrm{diff}} + \epsilon L^{\mathrm{con}}$.

\begin{figure}[t]
\centering
\includegraphics[width=\linewidth,trim={0cm 0cm 0cm 0cm}, clip]{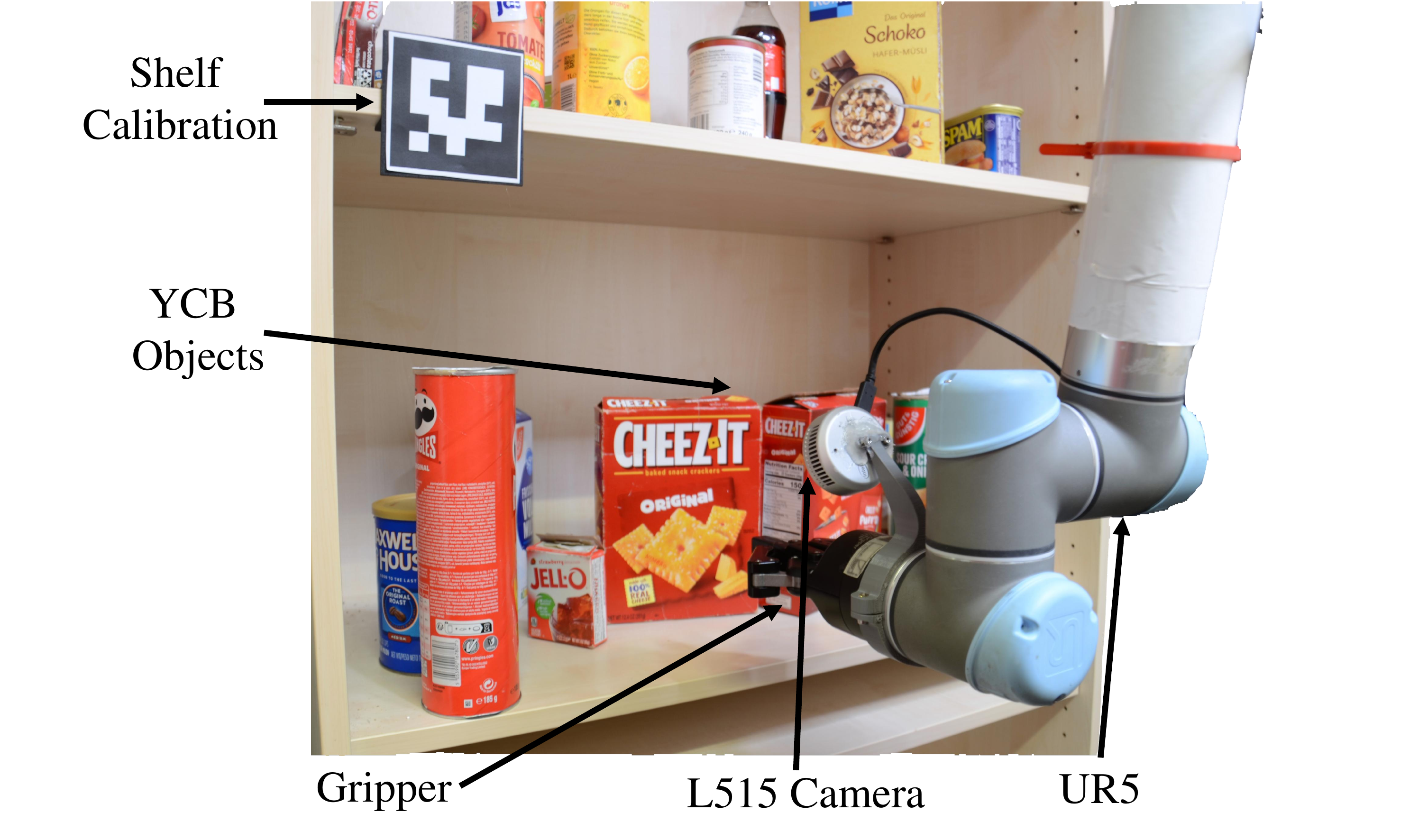}
\caption{Real-world environment showing a shelf scenario. The UR5 is equipped with an Robotiq parallel-jaw gripper and a Realsense L515
RGB-D camera to create a calibrated representation of the scene.}
\label{real_world_setup}
\vspace{-10px}
\end{figure}

\begin{figure*}[t]
\centering

\captionsetup[subfigure]{labelformat=empty}
\subfloat[\label{fig:sim_eval_non_pushing_label}]{\includegraphics[width=0.8\textwidth,trim={0cm 4.3cm 0cm 4.3cm},clip]{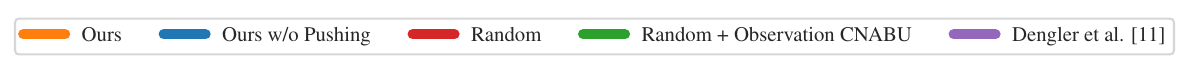}}  
\vspace{-0.6cm}

\subfloat[Low Occlusion - Occupancy\label{fig:sim_eval:a}]{\includegraphics[width=.24\textwidth,trim={0cm 0cm 0cm 0cm},clip]{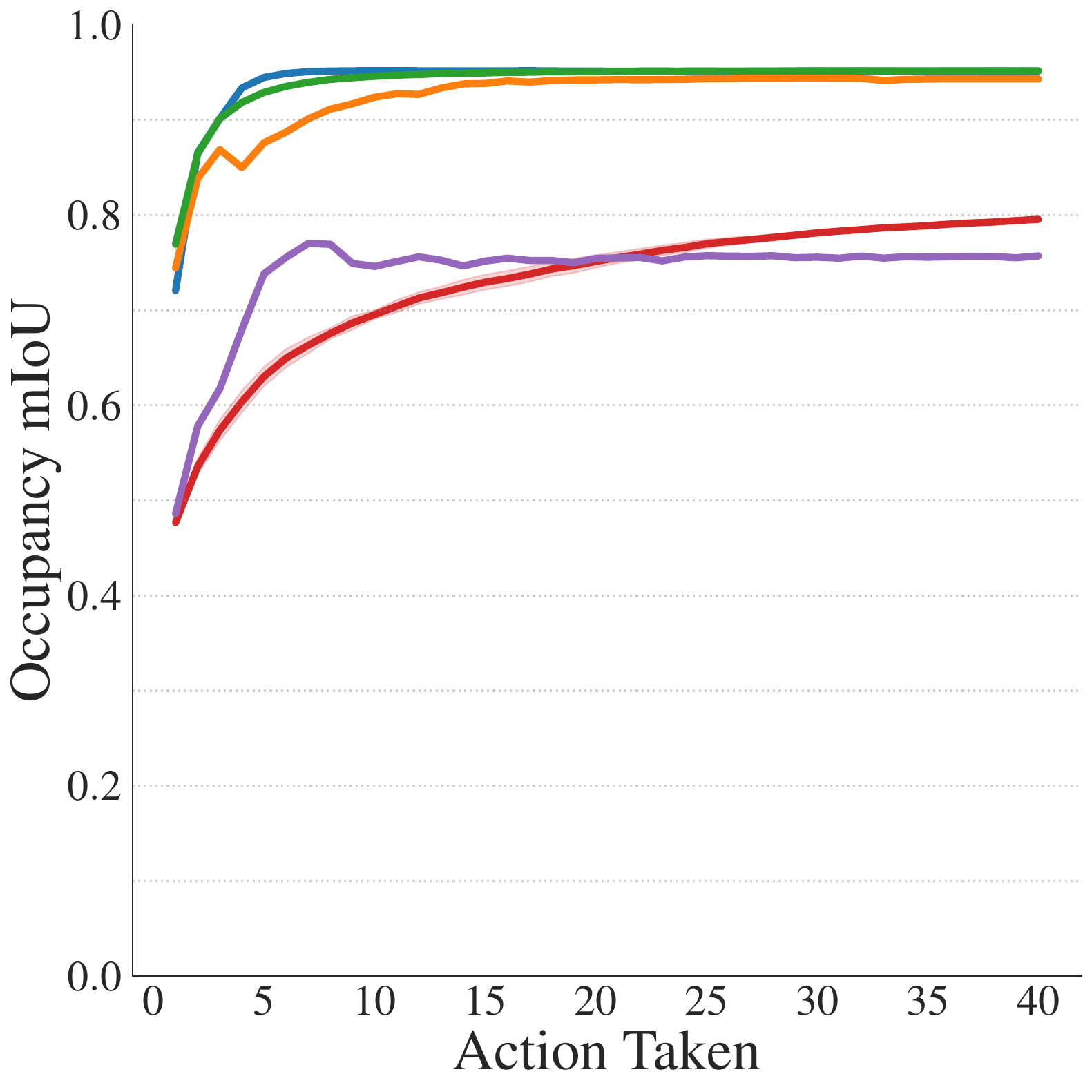}}
\hfill
\subfloat[High Occlusion - Occupancy\label{fig:sim_eval:b}]{\includegraphics[width=.24\textwidth, trim={0cm 0cm 0cm 0cm},clip]{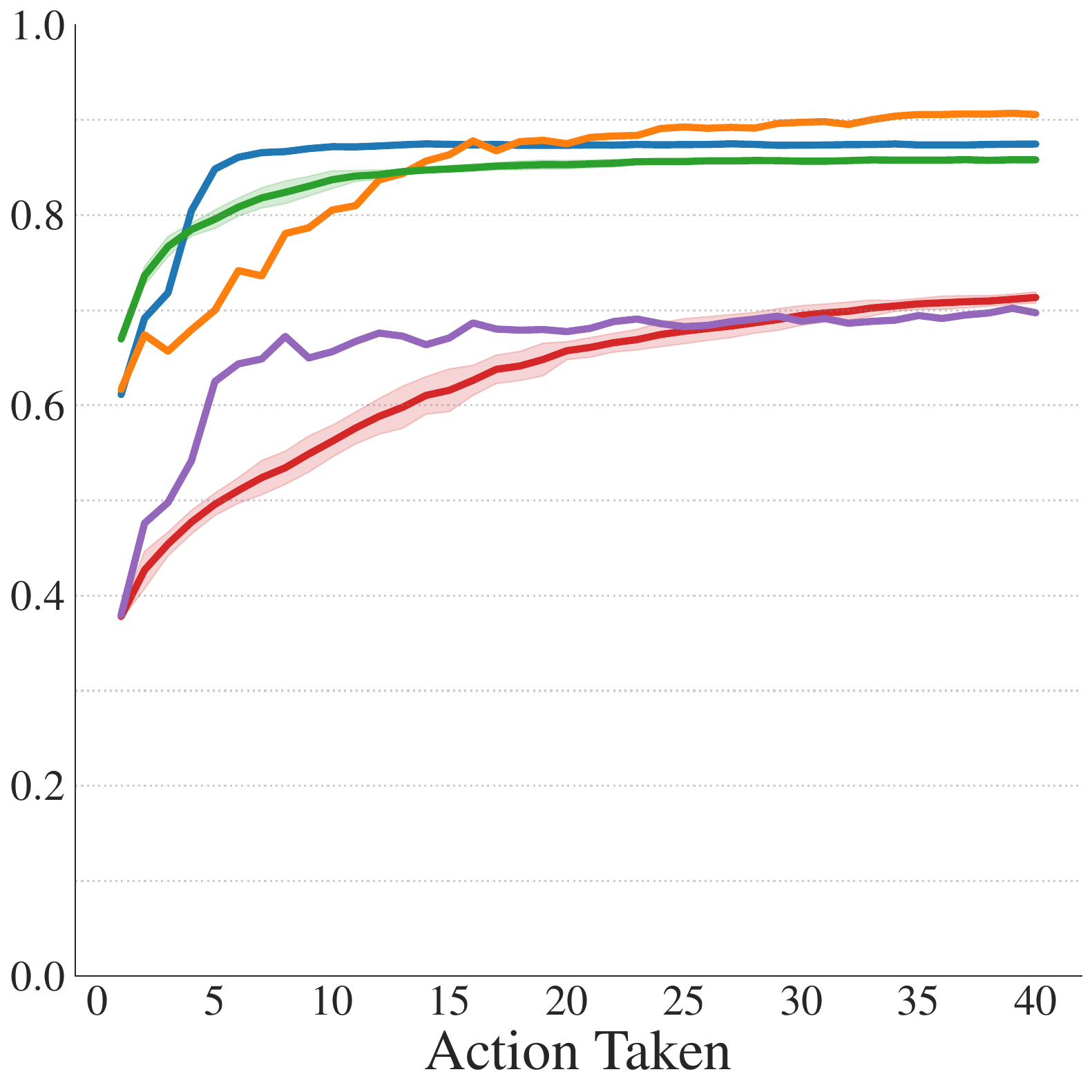}}
\hfill
\subfloat[Low Occlusion - Semantics\label{fig:sim_eval:c}]{\includegraphics[width=.24\textwidth, trim={0cm 0cm 0cm 0cm},clip]{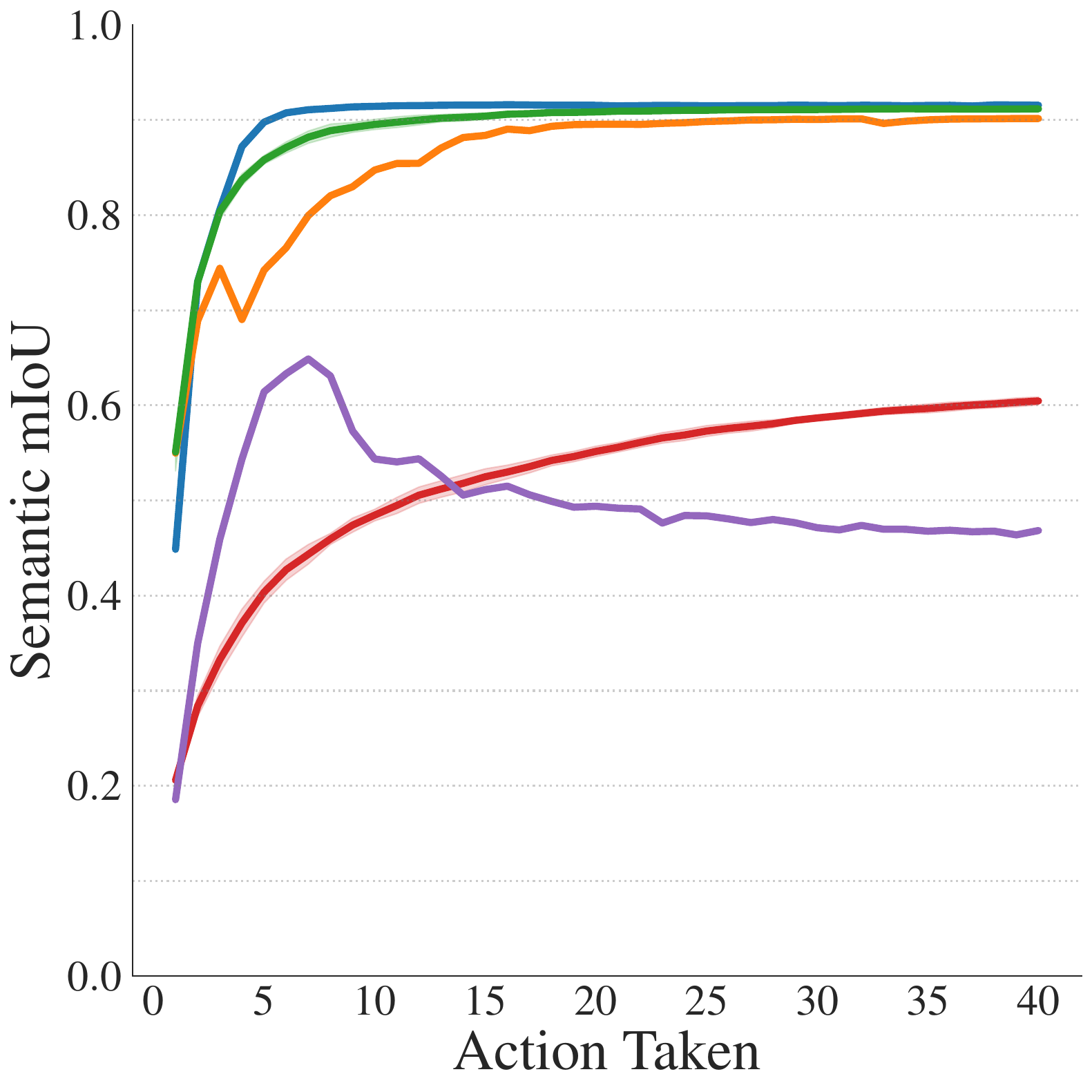}}
\hfill
\subfloat[High Occlusion - Semantics\label{fig:sim_eval:d}]{\includegraphics[width=.24\textwidth,trim={0cm 0cm 0cm 0cm},clip]{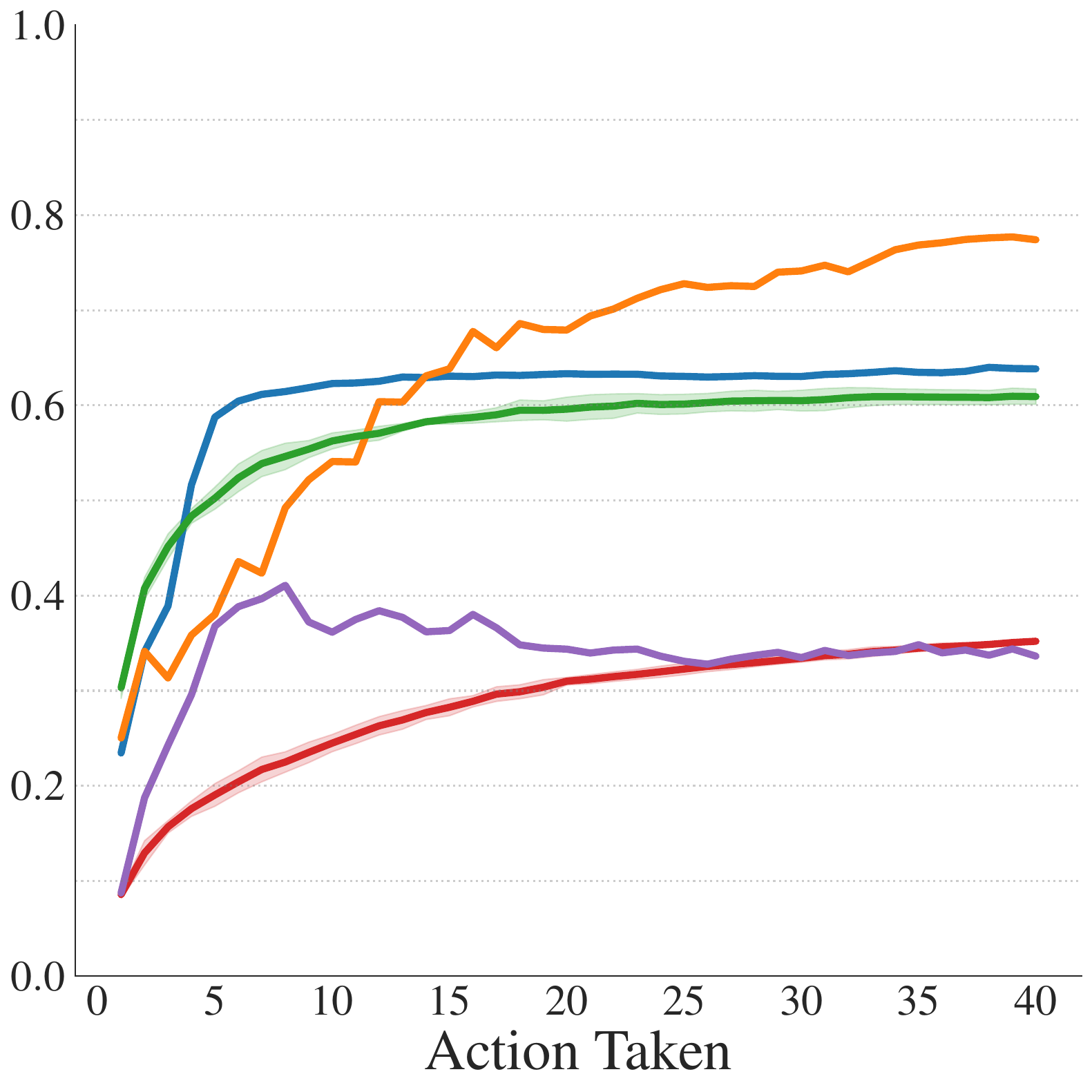}}  
\caption{
Simulation results for Manipulation Enhanced Mapping against SOTA and non-pushing baselines—showing both occupancy and semantic IoUs over time for each method. Our method outperforms all baselines in highly occluded scenes, while not having degraded performance in low occlusion scenes. Standard deviation of performance of random baselines over random seeds is represented as shading around each plot.
}
\label{fig:simulation eval}
\vspace{-10px}
\end{figure*}

\section{Experiments}
\label{sec:Experiments & Results}
We perform four core experiments to evaluate our approach. First, we test in simulation to highlight our pipeline's improvements in map completeness and accuracy compared to state-of-the-art~\cite{dengler2023viewpoint}. Next, we present a series of ablations of our method and evaluate several interactive baselines. We then present some experiments on the generalization of the trained CNABUs to within-class shape variations.
Finally, we study the robustness of our system in terms of its zero-shot transferability to a physical setup.
Further evaluations, which validate the individual CNABU's performance and the use of VIG as a reward proxy, are provided in Appendices  \ref{sec:cnabu_performance_evaluation} and \ref{sec:validating_vig}.

\subsection{Experimental Setup}
\label{eval_setup}

Our task setup consists of a shelf scene with a UR5 arm for observation and action execution (\autoref{real_world_setup}).
The robot is fixed to a table facing an occluded shelf and equipped with a Robotiq parallel-jaw gripper for manipulation and an RGB-D camera for observations. In simulation, the ground truth observations and segmentations are provided by rendering.

The real-world setup is similar, but with a few notable differences. 
For action execution, ROS and MoveIt \cite{Chitta2012MoveItTopics} are used. The depth image is obtained from an intel Realsense L515 camera and the semantic segmentation in the real world is performed using segment anything 2 (SAM2) \cite{ravi2024sam} and a strategy similar to LSeg \cite{li2022languagedriven,jatavallabhula2023conceptfusion}. We take detected masks from SAM2 and crop the original image around them. Next, we extract their CLIP \cite{radford2021learning} embeddings and compute their cosine similarity to the language embeddings of our target classes, whose prompts we list in Appendix~\autoref{sec:segmentation_prompts}. Finally, we normalize the similarity scores to classify each mask.

\begin{figure*}[!ht]
\centering
\captionsetup[subfigure]{labelformat=empty}
\subfloat[\label{fig:sim_eval_pushing_legend}]{\includegraphics[width=.8\textwidth,trim={0cm 4.3cm 0cm 4.3cm},clip]{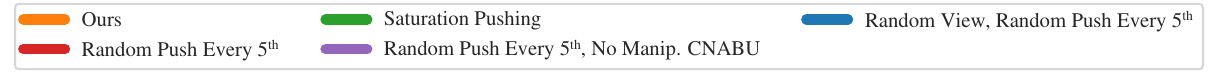}}  
\vspace{-10px}

\subfloat[Low Occlusion - Occupancy\label{fig:pushing_baselines_comparison:d}]{\includegraphics[width=.24\linewidth,trim={0cm 0cm 0cm 0cm},clip]{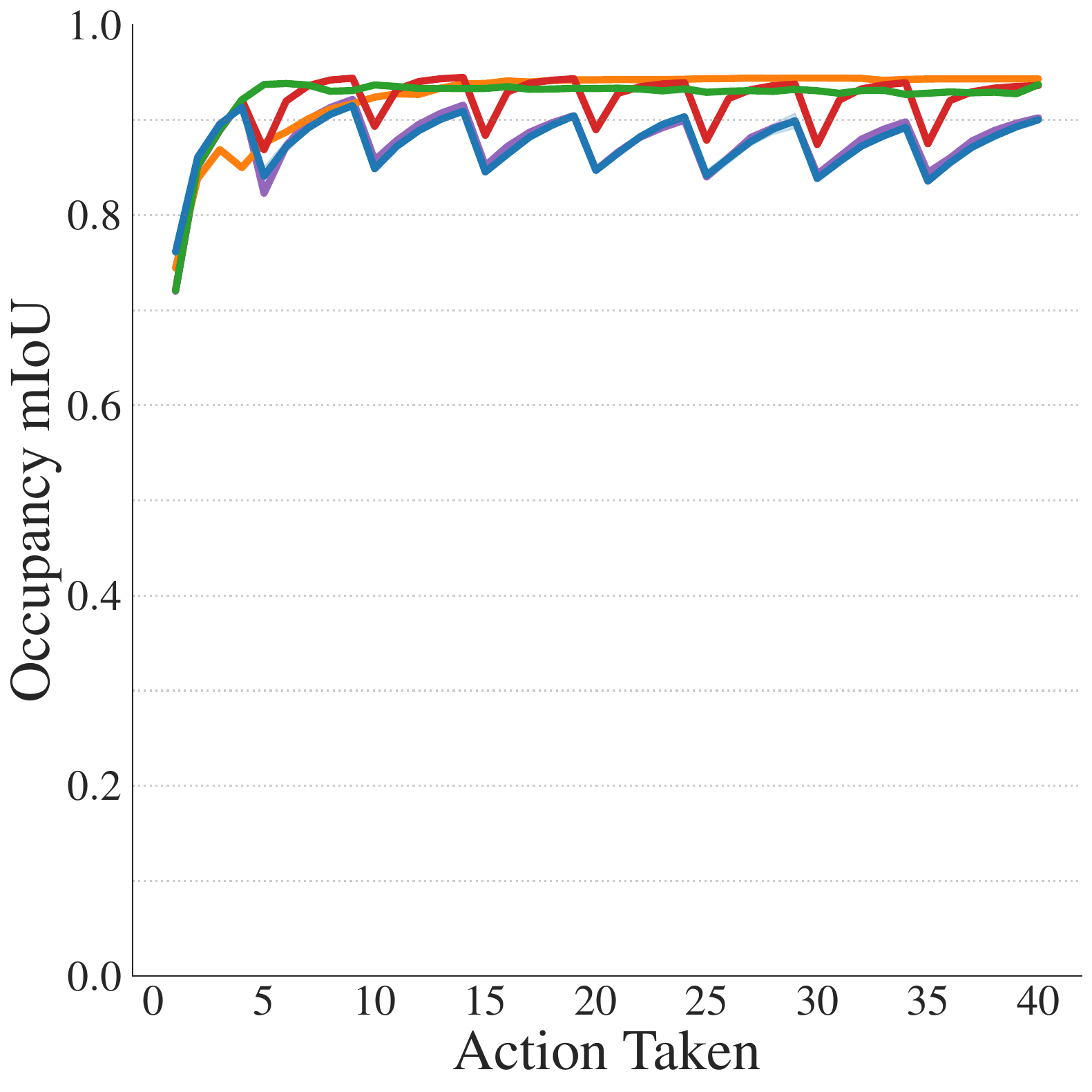}} 
\hfill
\subfloat[High Occlusion - Occupancy\label{fig:pushing_baselines_comparison:c}]{\includegraphics[width=.24\linewidth, trim={0cm 0cm 0cm 0cm},clip]{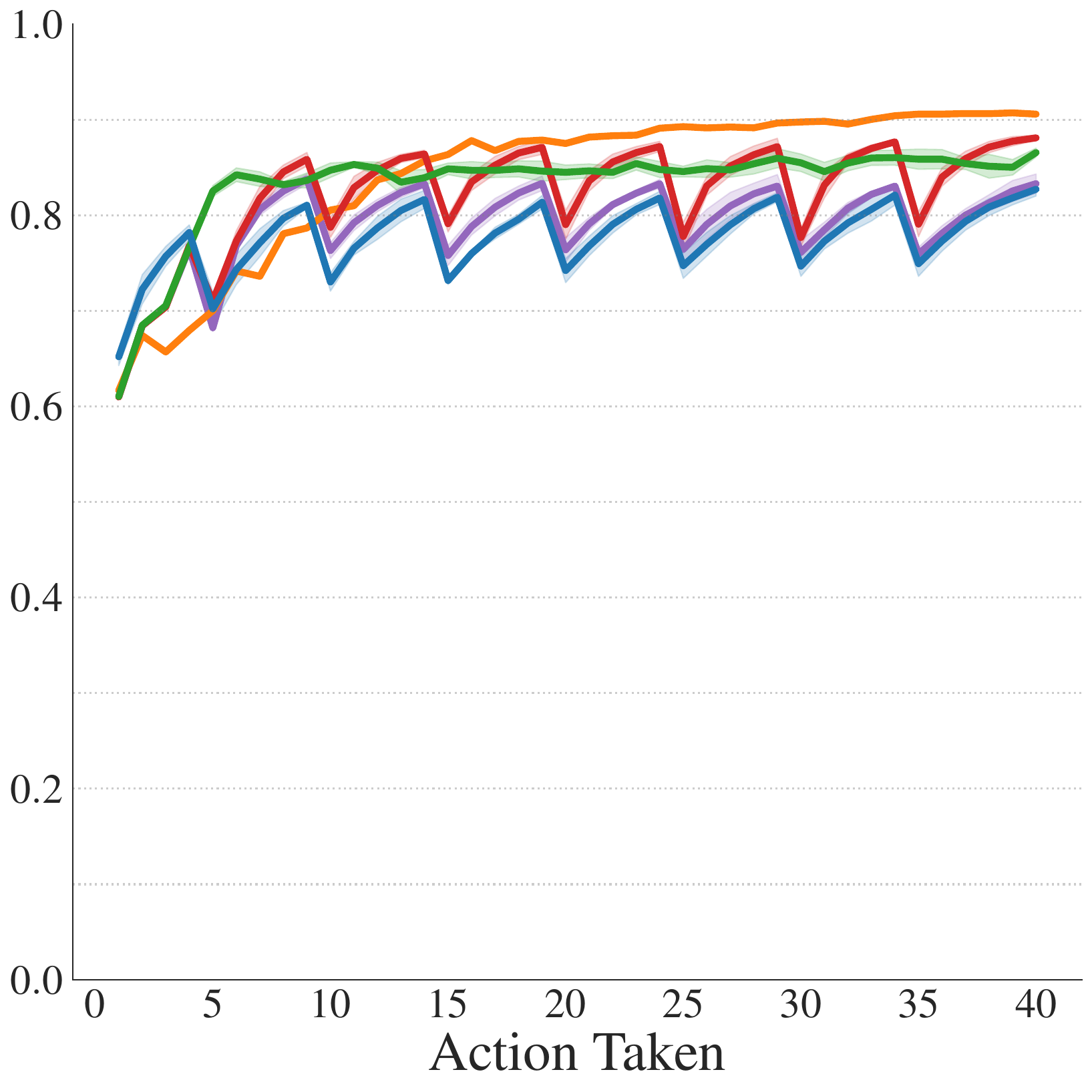}}
\hfill
\subfloat[Low Occlusion - Semantics\label{fig:pushing_baselines_comparison:b}]{\includegraphics[width=.24\linewidth,trim={0cm 0cm 0cm 0cm},clip]{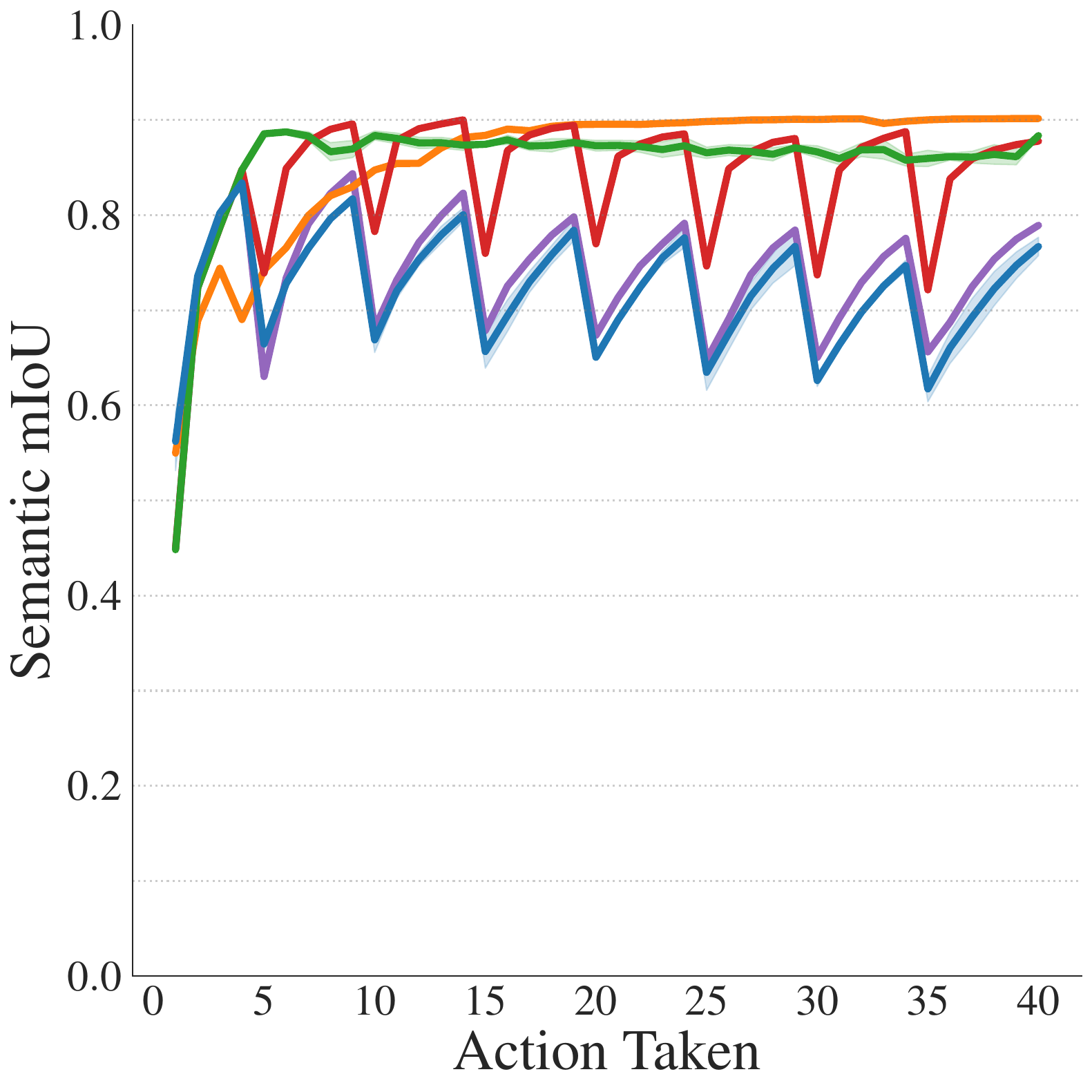}} 
\hfill
\subfloat[High Occlusion - Semantics\label{fig:pushing_baselines_comparison:a}]{\includegraphics[width=.24\linewidth, trim={0cm 0cm 0cm 0cm},clip]{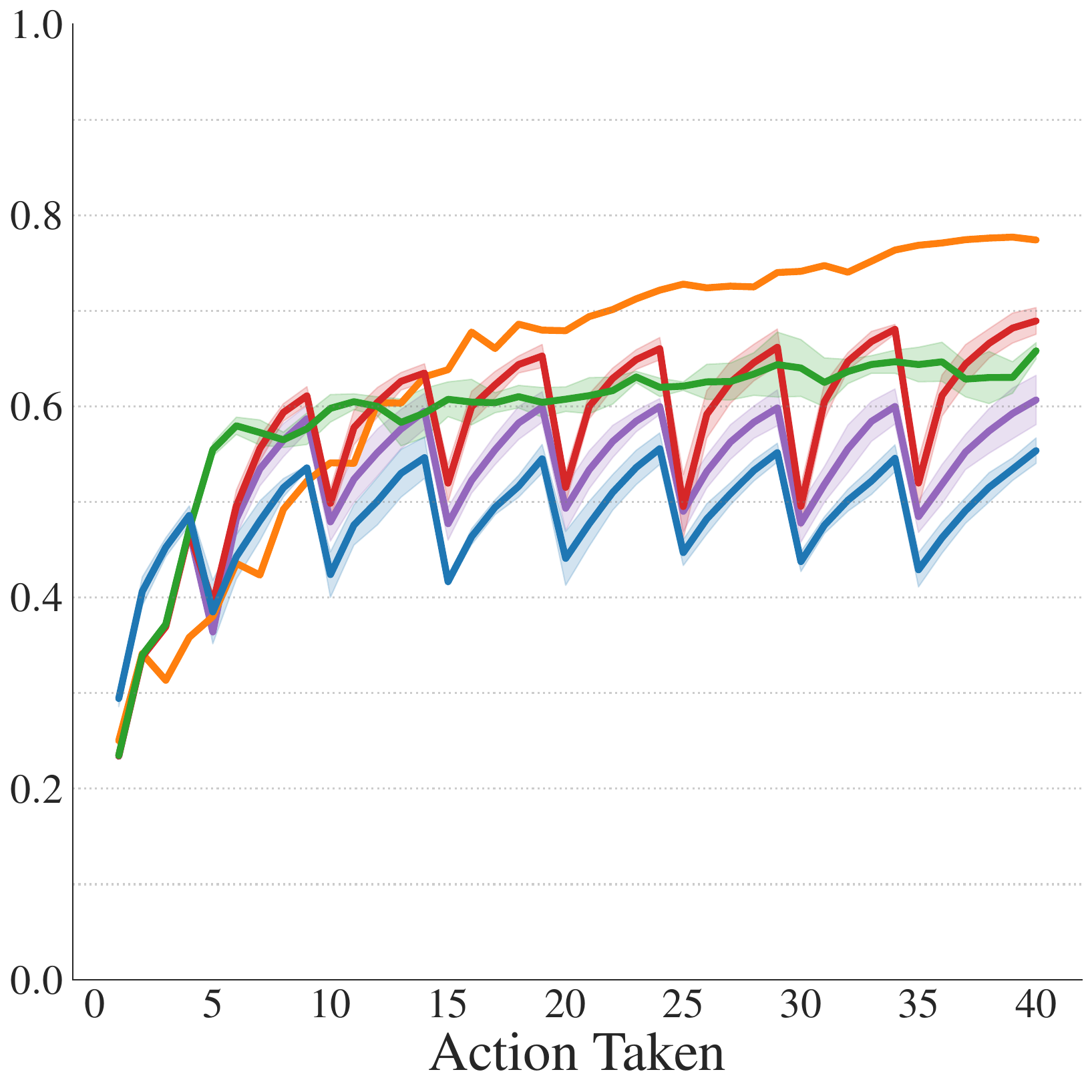}}
\caption{Simulation tests of push selection alternatives. Note how our method not only achieves better mIoUs than any of the other methods, it does so consistently across all the steps, avoiding uninformative or overly aggressive manipulation actions. Also, next-best-view with the observation belief prediction but without push belief prediction (purple) leads to degraded performance. Standard deviation of performance of random baselines is represented as shading around each plot.}
\label{fig:sim_eval_pushing_pushing}
\vspace{-10px}
\end{figure*}

\subsection{Simulation Experiments}

These experiments consider both low and high occlusion scenarios for manipulation-enhanced mapping. 
We generate 100 low occlusion scenarios via rejection sampling, using our sampling method described in Appendix~\ref{sec:dataset_generation}, but keeping only scenarios for which at least one object cannot be seen from any of the 300 viewpoints.
We then crafted 25 high occlusion scenarios by hand to be challengingly crowded and with many objects occluded.
We provide examples of each category in Figs. \ref{fig:simulated_example_hard} and \ref{fig:simulated_example_easy} in the Appendix. 
The robot begins with a naive uniform map prior. 


We compare our work (\textbf{Ours}) with the following baselines. 
First, we reimplemented the approach of \citet{dengler2023viewpoint} and fine-tuned the network weights provided by the authors for only 5,000 action steps. 
Although their experimental setup closely aligns with ours, our tests introduce a lower upper shelf board height and more densely sampled object configurations.
Second, the \textbf{Random} baseline randomly samples a set of unique views $[v^r_0, \dots, v^r_n] \in \mathds{V}$ and uses standard metric-semantic occupancy mapping \cite{thrun2005probabilistic}. 
We also combine random view selection with our observation belief predictor $\sigma_o$, \textbf{Random + Observation CNABU}.
We also compare an ablation of our pipeline that does not use manipulation, \textbf{Ours w/o pushing}.

Metric and semantic mIoU compared to the ground truth map at time $t$ are plotted in \autoref{fig:simulation eval}.
We observe that the previous S.O.T.A. for unstructured MEM, \citet{dengler2023viewpoint}, explores efficiently at early stages. However, with more of the scene uncovered and new areas being harder to observe, its performance degrades to comparable or worse than random, particularly in terms of semantic mIoU. 
We attribute this performance degradation to two key challenges: the use of heuristic-based action switching and the lack of map updates after manipulation. 
The heuristic switching mechanism relies on hand-crafted rules to alternate between observation and manipulation actions, and may not always select the most informative action, occasionally choosing to push too early, too often, or to continue observing when manipulation would be more beneficial. 
Our POMDP-based action selection overcomes this problem and does not select manipulation actions when they are not beneficial for higher information gain.
Furthermore, because the \citet{dengler2023viewpoint}'s pipeline does not update its belief after a push, it requires multiple subsequent observations to reconcile inconsistencies between the actual scene and the previously assumed map representation. This delay in belief correction leads to inefficient re-exploration and degraded semantic accuracy, as the agent lacks a reliable signal for where to focus its attention.

Moreover, we observe that belief prediction is a powerful approach, leading to excellent scene coverage in low occlusion scenes even without pushing. 
In highly occluded scenarios, pushing is required to make progress after the visible surfaces are observed. Our method uses pushing to achieve significantly higher mIoUs. 
Note that its IoU growth is slower early on, because pushing does not provide information until a viewpoint step is taken in the following action.

\subsection{Push Selection Alternatives}
Next, we validate our POMDP push selection strategy. Unless otherwise noted, the same belief prediction networks are provided to each method. 
Three of the strategies push regularly at a five-step interval, which is a typical rate at which observation actions provide diminishing returns. 
The first, \textbf{Random Push Every $5^{th}$}, plans views using NBV and $\sigma_o$, and randomly samples a push every five steps and updates its belief using $\sigma_m$. The second, \textbf{Random Push Every $5^{th}$, No Manip. CNABU} is the same, but does not use $\sigma_m$ to update its belief after pushing. The third, \textbf{Random View Random Push Every $5^{th}$}, chooses random views and randomly pushes every five steps without using $\sigma_m$. 
Finally, we consider a heuristic that performs a random push when VIG seems to saturate, \textbf{Saturation Pushing}. The saturation threshold is when two consecutive estimates of VIG during NBV differ by less than 2\%. Saturation pushing uses $\sigma_m$ after each push to update its belief. A comparison of selection strategies is given in \autoref{sec:baselines_table} in the Appendix.

Results are shown in \autoref{fig:sim_eval_pushing_pushing}. Each baseline that has a random component is run three times with different random seeds. We observe that push belief prediction is beneficial to manipulation-enhanced mapping, even when random pushes are being executed. The blue and purple curves show methods that are not informed by $\sigma_m$.
Further, we see that the saturation pushing (green) does not observe post-push performance drops, but it is ultimately outperformed by random informed pushes and our method. Overall, our method still achieves the best performance, most strikingly in highly occluded scenes.

\vspace{-3px}
\begin{figure*}[!t]
\centering
\includegraphics[width=1 \linewidth]{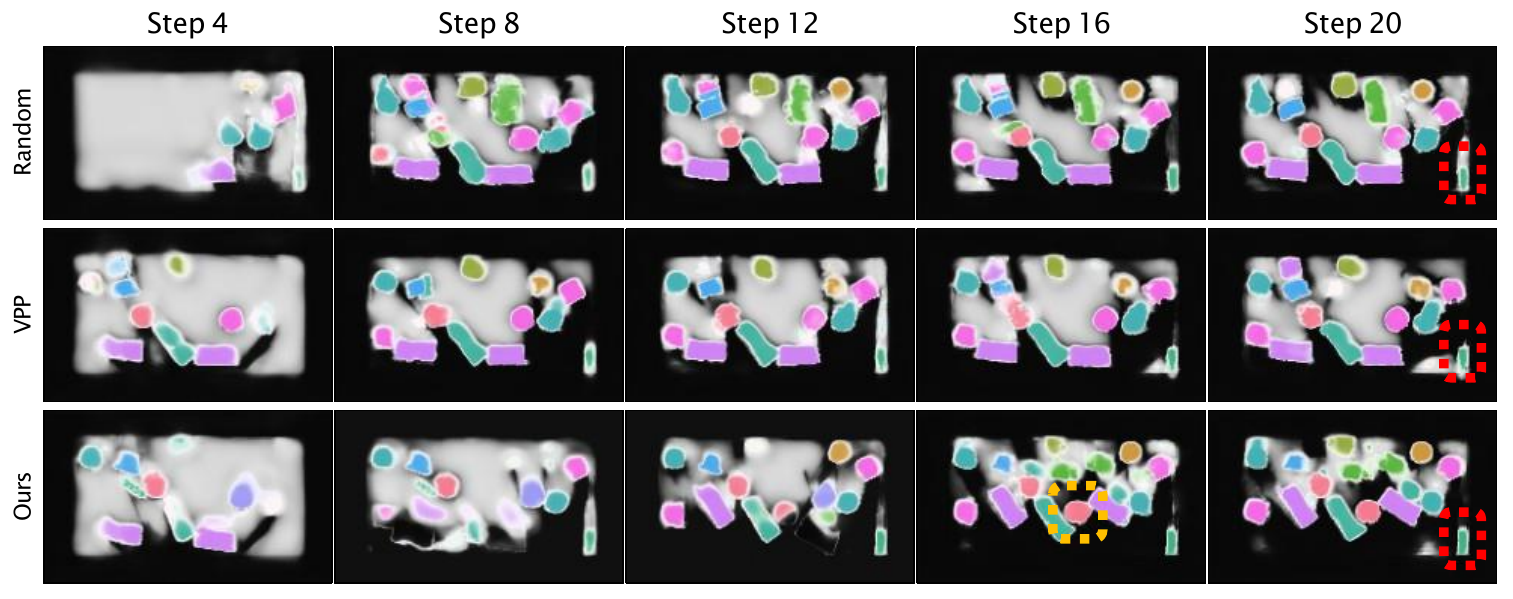}
\\
\vspace{5px}
\captionsetup[subfigure]{labelformat=empty}
\subfloat[
]{\includegraphics[width=1\textwidth,trim={0cm 0cm 0cm 0cm},clip]{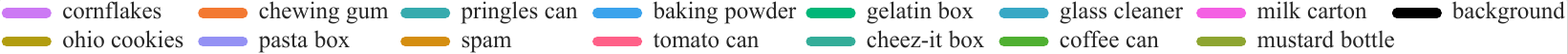}}  
\vspace{-10px}
\caption{Qualitative real-world experiment results. Note how VPP and Random baselines are unable to fully explore the environment due to its occlusions, while our method is able to better explore it via reasonable manipulations. In Step 16, Ours, we highlight in yellow one of the scene objects revealed by manipulation. We highlight in red a persistent hallucination across all 3 methods, likely due to unreliable semantic segmentation and significant camera noise at that corner of the shelf .}
\vspace{-10px}
\label{fig:Qualitative Comparison}
\end{figure*}
\subsection{Out of Distribution Shape Experiments}

\begin{figure}[htbp!]
    \centering

    \includegraphics[width=0.5\textwidth,trim={0cm 4.9cm 0cm 4.9cm},clip]{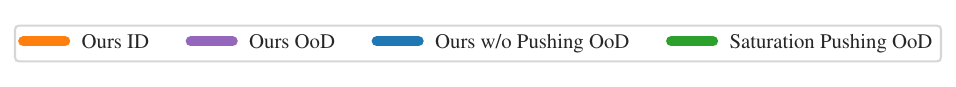}

    \begin{subfigure}{0.24\textwidth}
        \includegraphics[width=\linewidth,trim={0.2cm 0.2cm 0cm 0cm},clip]{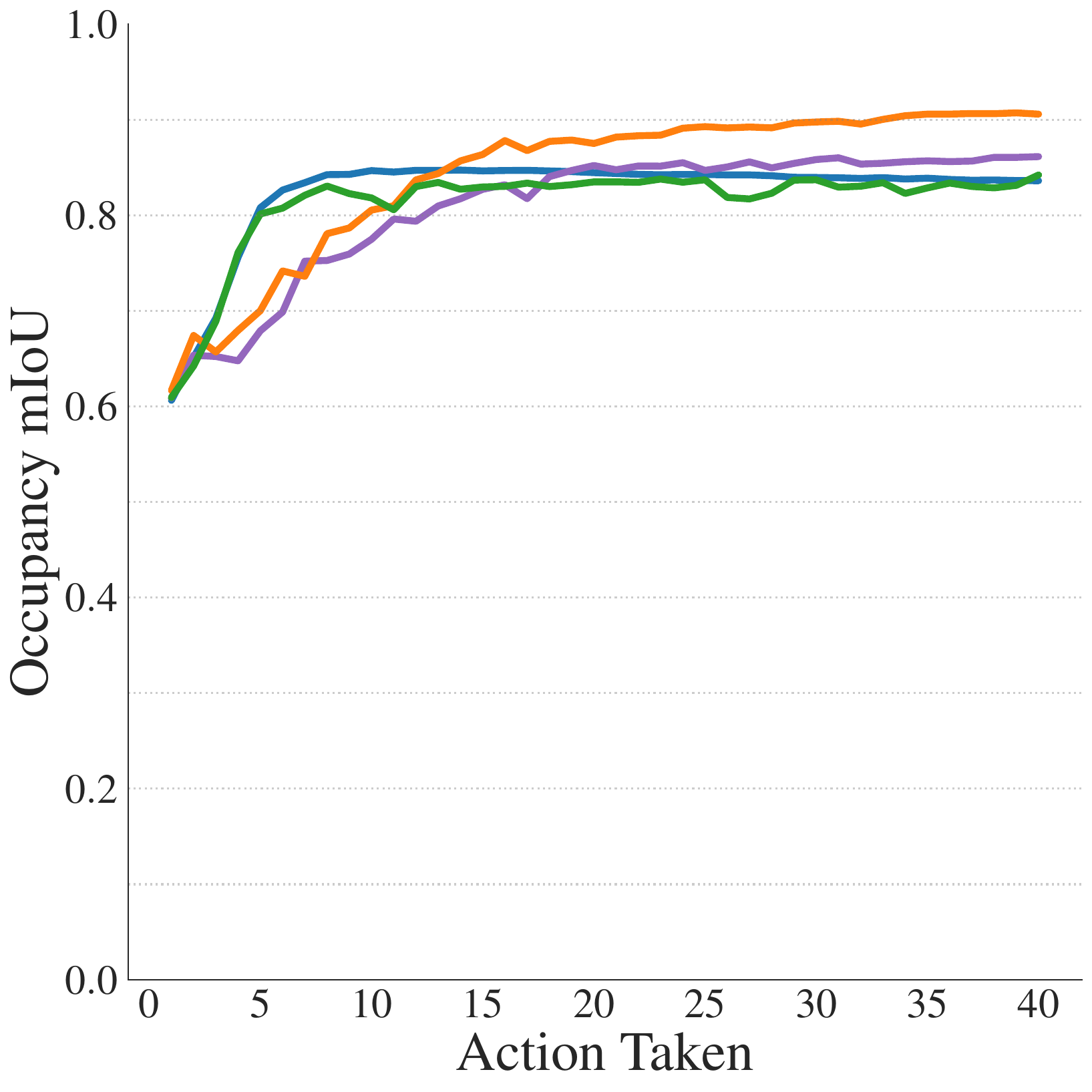}
        \caption{Occupancy Results (IoU)}
    \end{subfigure}
    \hfill
    \begin{subfigure}{0.24\textwidth}
        \includegraphics[width=\linewidth,trim={0.2cm 0.2cm 0cm 0cm},clip]{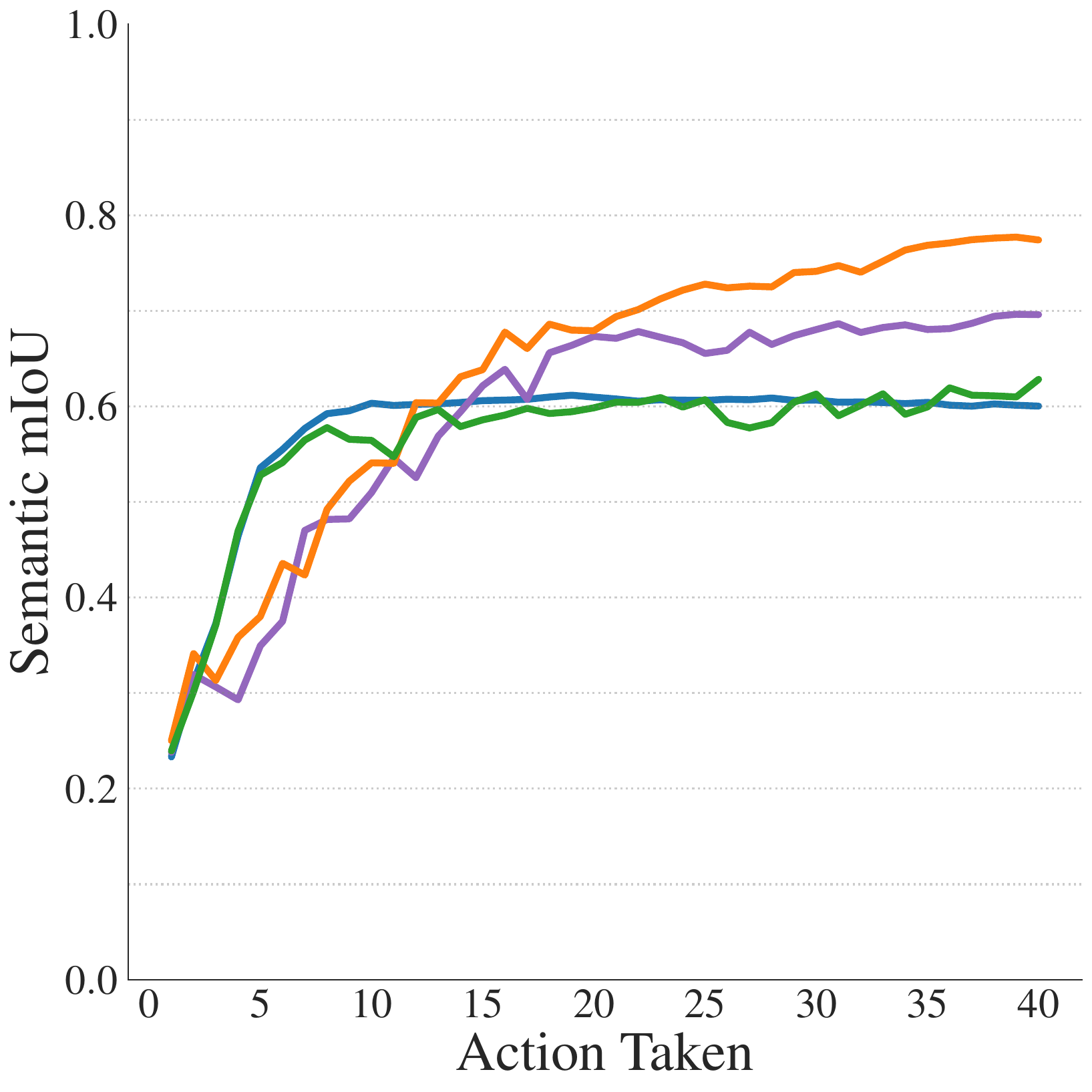}
        \caption{Semantic Results (mIoU)}
    \end{subfigure}

    \caption{OoD Results in \textit{High Occlusion} scenes.}
    \label{fig:OOD_results}
    \vspace{-5px}
\end{figure}

In order to evaluate the robustness of the proposed pipeline to shape variations within a given class, we modify the simulation setup. For each individual instance of an object in a given scene, we randomly and independently rescale the object's mesh on each of its principal axis by a factor chosen uniformly at random within $[0.8,1.2]$, and re-run our simulation experiments in the high-occlusion scenes with these out-of-distribution shapes for the strongest baselines in simulation. As seen in \autoref{fig:OOD_results}, we observe a decrease in the agent's performance when out of distribution, but note that in the long run the performance remains consistent with our ID experiments. This is notable since we do not perform any sort of shape augmentation during the training of the CNABUs, suggesting that the fact that they are grounded in observed maps helps them overcome some level of distribution shift. Naturally, we presume that adding shape augmentation and more diverse object geometries of the same class during training should further help reduce this OOD performance gap. 

\subsection{Hardware Experiments}
We also performed 10 real-world experimental runs on a UR5 as described in Sec. \ref{eval_setup}.
All results are collected in a zero-shot fashion, i.e., no fine-tuning on real data was performed.
We set the budget to 20 steps and sampled a fixed set of 75 reachable camera poses in front of the shelf for $\mathds{V}$. 

We handcrafted 10 challenging scenes, each with an average of 18 objects from the YCB dataset \cite{calli2015benchmarking}, where pushing is required to reveal other objects.
We collect the ground truths by removing the top of the shelf at the end of each episode to manually score the final maps. 
We score each scenario according to the status of \textbf{all of the objects} present in the map. 
Each object in the map is classified in four categories: 1) \textbf{Correctly Found} if the majority of the object is correctly represented in the map with the right class; 2) \textbf{Misclassified But Found} if the majority of the object is present in the occupancy map but is mislabeled; 3) \textbf{Not Found} - if the majority of the object is absent from the occupancy map and 4) \textbf{Hallucinated} if an object that is not present in the scene is present in the map.
Due to the complexity of precisely resetting a scene multiple times for different baselines and the fact that both random + observation CNABU and our approach without pushing are about as strong as other pushing baselines (see Figs. \ref{fig:simulation eval} and \ref{fig:sim_eval_pushing_pushing}), we only compare against them in the physical experiments. For each model, we report the total quantity of each detection at time step 20 summed over all 10 trials. 

Results in \autoref{tab:quantitative_real_experiments} show that with zero-shot transfer from sim-to-real, the proposed method still manages to retain its edge over the compared baselines. Note that all methods compared use calibrated belief prediction. However, little difference is seen between viewpoint planning (Ours w/o Pushing) and random viewpoint choices. We expect that this is due to a domain gap caused by camera noise from the realsense L515 leading to some strong artifacting in the depth images and the inaccuracies of the open-set semantic segmentation pipeline. We found the SAM2 segmenter performed particularly poorly for oblique angles or partial object views, resulting in many missed instance detections and misclassifications. For instance, there is a consistent artifact on the segmentation pipeline, which consistently classifies the bottom right corner of the shelf as a gelatin box, as seen in all methods in \autoref{fig:Qualitative Comparison}. Nevertheless, we can see that our methods (both with pushing and without pushing) greatly reduce the number of hallucinations and improve the number of correctly identified objects. Further, our complete pipeline reveals 39\% of the objects that were previously unseen by the non-interactive baselines, performance consistent with the simulation experiments, despite the significant sim-to-real gap, particularly in segmentation performance, leading to many of the newfound objects being incorrectly classified.

\begin{table}[tbhp]
  \centering
  \caption{Comparing our method to the strongest baselines in zero-shot transfer to real-world shelves. Total counts of each type of detection are reported across our 10 trials.}
  \resizebox{0.48\textwidth}{!}{%
  \begin{tabular}{lcccc}
  \hline
    \textbf{Policy} & \textbf{\makecell{Correctly\\Found $\uparrow$}} & \textbf{\makecell{Missclassified\\But Found$\uparrow$}} & \textbf{\makecell{Not Found$\downarrow$}} & \textbf{Hallucinated$\downarrow$} \\ \hline
    \textbf{Random} & 72 & 47 & 52 & 11\\ 
    \textbf{Ours w/o Pushing} & 81 & 38 & 52 & \textbf{6} \\ 
    \textbf{Ours} & \textbf{85} & \textbf{52} & \textbf{35} & 7 \\ \hline

  \end{tabular}%
  }
  \label{tab:quantitative_real_experiments}
\vspace{-10px}

\end{table}

\section{Limitations} 
\label{sec:limitations}

Limitations of our method include the need for representative simulation training data or ground truth segmented maps. It also relies on high-quality semantic segmentation, and although the computer vision field is making significant progress on segmentation, segmentation accuracy is still too low for many robotics applications in occluded, poorly lit and partial views, especially in open-set scenarios.

Computation times for our POMDP solver vary but take on the order of several seconds, due to the need for information gain calculations and belief propagation for many actions. Our current framework na\"{i}vely samples manipulation actions during action selection, and more intelligent action sampling could improve computational efficiency. This will be especially important when including additional manipulation actions, e.g., grasping.

There is currently a significant sim-to-real gap that could be addressed by fine-tuning on real data or performing domain randomization of the object dynamics during data collection in simulation to help improve real-world performance. Further, our maps are defined over dense voxel grids, which poses scalability challenges when applied to larger spaces. Moreover, the mapped region is a fixed volume and we assume prior knowledge about the fixed parts of the environment (e.g., shelf) for motion planning. These assumptions should be relaxed to address fully unstructured and unknown worlds.
Finally, our maps use a closed-world semantic labeling, and extending belief propagation to open-world segmentation would be an interesting frontier to explore. 

\section{Conclusion}


This paper presented a POMDP approach to the manipulation-enhanced mapping problem in which the solver decides between changing a camera view and manipulating objects to map an area cluttered with objects. It relies on the novel Calibrated Neural-Accelerated Belief Update map-space belief propagation approach, which allows a unified treatment of both viewpoint change and manipulation actions. Using well-calibrated beliefs allows the POMDP solver to make decisions between different action types according to the most informative outcome.
Experimental results in both simulation and on a real robot show that our approach outperforms non-manipulating and heuristic manipulation baselines in terms of occupancy and semantic map accuracy.
Overall, this work offers a promising new framework for robots navigating and manipulating real-world cluttered environments. 

\section{Acknowledgments}

This work used the NCSA Delta GPU cluster through allocation CIS240410 from the Advanced Cyberinfrastructure Coordination Ecosystem: Services \& Support (ACCESS) program \cite{access_citation}, which is supported by U.S. National Science Foundation grants \#2138259, \#2138286, \#2138307, \#2137603, and \#2138296. This work has partly been supported by the European Commission under grant agreement numbers 964854 (RePAIR) and by the BMBF within the Robotics Institute Germany, grant \#16ME0999. This work was also partially supported by NIFA/USDA Awards \#2020-67021-32799 
and \#2021-67021-34418.

\bibliographystyle{plainnat}
\bibliography{other_references,joao_references}

\begin{thebibliography}{47}
\providecommand{\natexlab}[1]{#1}
\providecommand{\url}[1]{\texttt{#1}}
\expandafter\ifx\csname urlstyle\endcsname\relax
  \providecommand{\doi}[1]{doi: #1}\else
  \providecommand{\doi}{doi: \begingroup \urlstyle{rm}\Url}\fi

\bibitem[Bircher et~al.(2016)Bircher, Kamel, Alexis, Burri, Oettershagen, Omari, Mantel, and Siegwart]{bircher2016three}
Andreas Bircher, Mina Kamel, Kostas Alexis, Michael Burri, Philipp Oettershagen, Sammy Omari, Thomas Mantel, and Roland Siegwart.
\newblock {Three-dimensional coverage path planning via viewpoint resampling and tour optimization for aerial robots}.
\newblock \emph{Autonomous Robots}, 40\penalty0 (6):\penalty0 1059--1078, 2016.

\bibitem[Boerner et~al.(2023)Boerner, Deems, Furlani, Knuth, and Towns]{access_citation}
Timothy~J. Boerner, Stephen Deems, Thomas~R. Furlani, Shelley~L. Knuth, and John Towns.
\newblock {ACCESS}: Advancing innovation: {NSF}’s advanced cyberinfrastructure coordination ecosystem: Services \& support.
\newblock In \emph{Practice and Experience in Advanced Research Computing 2023: Computing for the Common Good}, PEARC '23, page 173–176, New York, NY, USA, 2023. Association for Computing Machinery.
\newblock ISBN 9781450399852.
\newblock \doi{10.1145/3569951.3597559}.
\newblock URL \url{https://doi.org/10.1145/3569951.3597559}.

\bibitem[Bohg et~al.(2017)Bohg, Hausman, Sankaran, Brock, Kragic, Schaal, and Sukhatme]{bohg2017interactive}
Jeannette Bohg, Karol Hausman, Bharath Sankaran, Oliver Brock, Danica Kragic, Stefan Schaal, and Gaurav~S Sukhatme.
\newblock Interactive perception: Leveraging action in perception and perception in action.
\newblock \emph{IEEE Transactions on Robotics}, 2017.
\newblock \doi{10.1109/TRO.2017.2721939}.
\newblock URL \url{https://ieeexplore.ieee.org/document/8007233}.

\bibitem[Calli et~al.(2015)Calli, Walsman, Singh, Srinivasa, Abbeel, and Dollar]{calli2015benchmarking}
Berk Calli, Aaron Walsman, Arjun Singh, Siddhartha Srinivasa, Pieter Abbeel, and Aaron~M Dollar.
\newblock Benchmarking in manipulation research: The {YCB} object and model set and benchmarking protocols.
\newblock \emph{IEEE Robotics and Automation Magazine}, 2015.

\bibitem[Chase~Kew et~al.(2021)Chase~Kew, Ichter, Bandari, Lee, and Faust]{10.1007/978-3-030-66723-8_5}
J~Chase~Kew, Brian Ichter, Maryam Bandari, Tsang-Wei~Edward Lee, and Aleksandra Faust.
\newblock {Neural Collision Clearance Estimator for Batched Motion Planning}.
\newblock In Steven~M LaValle, Ming Lin, Timo Ojala, Dylan Shell, and Jingjin Yu, editors, \emph{Algorithmic Foundations of Robotics XIV}, pages 73--89, Cham, 2021. Springer International Publishing.
\newblock ISBN 978-3-030-66723-8.

\bibitem[Chitta et~al.(2012)Chitta, Sucan, and Cousins]{Chitta2012MoveItTopics}
Sachin Chitta, Ioan Sucan, and Steve Cousins.
\newblock {MoveIt! [ROS Topics]}.
\newblock \emph{IEEE Robotics {\&} Automation Magazine}, 19\penalty0 (1):\penalty0 18--19, 3 2012.
\newblock ISSN 1070-9932.
\newblock \doi{10.1109/MRA.2011.2181749}.
\newblock URL \url{http://ieeexplore.ieee.org/document/6174325/}.

\bibitem[Choudhury et~al.(2017)Choudhury, Kapoor, Ranade, and Dey]{7989112}
Sanjiban Choudhury, Ashish Kapoor, Gireeja Ranade, and Debadeepta Dey.
\newblock {Learning to gather information via imitation}.
\newblock In \emph{2017 IEEE International Conference on Robotics and Automation (ICRA)}, pages 908--915, 2017.
\newblock \doi{10.1109/ICRA.2017.7989112}.

\bibitem[Coumans and Bai(2016--2021)]{coumans2021}
Erwin Coumans and Yunfei Bai.
\newblock {PyBullet}, a {Python} module for physics simulation for games, robotics and machine learning.
\newblock \url{http://pybullet.org}, 2016--2021.

\bibitem[Danielczuk et~al.(2019)Danielczuk, Kurenkov, Balakrishna, Matl, Wang, Mart{\'\i}n-Mart{\'\i}n, Garg, Savarese, and Goldberg]{danielczuk2019mechanical}
Michael Danielczuk, Andrey Kurenkov, Ashwin Balakrishna, Matthew Matl, David Wang, Roberto Mart{\'\i}n-Mart{\'\i}n, Animesh Garg, Silvio Savarese, and Ken Goldberg.
\newblock Mechanical search: Multi-step retrieval of a target object occluded by clutter.
\newblock In \emph{Proc.~of the IEEE Intl.~Conf.~on Robotics \& Automation (ICRA)}, 2019.
\newblock \doi{10.1109/ICRA.2019.8794143.}
\newblock URL \url{https://ieeexplore.ieee.org/document/8794143}.

\bibitem[Delmerico et~al.(2018)Delmerico, Isler, Sabzevari, and Scaramuzza]{Delmerico2018AReconstruction}
Jeffrey Delmerico, Stefan Isler, Reza Sabzevari, and Davide Scaramuzza.
\newblock {A comparison of volumetric information gain metrics for active 3D object reconstruction}.
\newblock \emph{Autonomous Robots}, 42\penalty0 (2):\penalty0 197--208, 2018.
\newblock ISSN 1573-7527.
\newblock \doi{10.1007/s10514-017-9634-0}.
\newblock URL \url{https://doi.org/10.1007/s10514-017-9634-0}.

\bibitem[Dengler et~al.(2023)Dengler, Pan, Kalagaturu, Menon, Dawood, and Bennewitz]{dengler2023viewpoint}
Nils Dengler, Sicong Pan, Vamsi Kalagaturu, Rohit Menon, Murad Dawood, and Maren Bennewitz.
\newblock Viewpoint push planning for mapping of unknown confined spaces.
\newblock In \emph{2023 IEEE/RSJ International Conference on Intelligent Robots and Systems (IROS)}, pages 1178--1184. IEEE, 2023.

\bibitem[Georgakis et~al.(2022)Georgakis, Bucher, Schmeckpeper, Singh, and Daniilidis]{georgakis2022learning}
Georgios Georgakis, Bernadette Bucher, Karl Schmeckpeper, Siddharth Singh, and Kostas Daniilidis.
\newblock {Learning to Map for Active Semantic Goal Navigation}.
\newblock In \emph{International Conference on Learning Representations}, 2022.
\newblock \doi{10.48550/arXiv.2106.15648}.
\newblock URL \url{https://openreview.net/forum?id=swrMQttr6wN}.

\bibitem[Gildardo and Latombe(2003)]{Gildardo2003AChecking}
Sanchez Gildardo and Jean-Claude Latombe.
\newblock A single-query bi-directional probabilistic roadmap planner with lazy collision checking.
\newblock In {Jarvis Raymond Austin} and Alexander Zelinsky, editors, \emph{Robotics Research}, pages 403--417, Berlin, Heidelberg, 2003. Springer Berlin Heidelberg.
\newblock ISBN 978-3-540-36460-3.

\bibitem[Hafner et~al.(2020)Hafner, Lillicrap, Ba, and Norouzi]{Hafner2020Dream}
Danijar Hafner, Timothy Lillicrap, Jimmy Ba, and Mohammad Norouzi.
\newblock Dream to control: Learning behaviors by latent imagination.
\newblock \emph{arXiv preprint arXiv:1912.01603}, 2020.

\bibitem[Hauser(2013)]{hauser13klampt}
K.~Hauser.
\newblock Robust contact generation for robot simulation with unstructured meshes.
\newblock In \emph{Proc.~of the Intl.~Symposium~on Robotic Research (ISRR)}, 2013.

\bibitem[Hepp et~al.(2018)Hepp, Dey, Sinha, Kapoor, Joshi, and Hilliges]{hepp2018learn}
Benjamin Hepp, Debadeepta Dey, Sudipta~N Sinha, Ashish Kapoor, Neel Joshi, and Otmar Hilliges.
\newblock Learn-to-score: Efficient 3d scene exploration by predicting view utility.
\newblock In \emph{Proceedings of the European conference on computer vision (ECCV)}, pages 437--452, 2018.
\newblock URL \url{https://link.springer.com/chapter/10.1007/978-3-030-01267-0_27}.

\bibitem[Hu et~al.(2024)Hu, Pan, Jin, Popovi{\'c}, and Bennewitz]{Hu24icra}
H.~Hu, S.~Pan, L.~Jin, M.~Popovi{\'c}, and M.~Bennewitz.
\newblock Active implicit reconstruction using one-shot view planning.
\newblock In \emph{Proc.~of the IEEE Intl.~Conf.~on Robotics \& Automation (ICRA)}, 2024.
\newblock \doi{10.1109/ICRA57147.2024.10611542.}
\newblock URL \url{https://ieeexplore.ieee.org/document/10611542}.

\bibitem[Huang et~al.(2021)Huang, Dominguez-Kuhne, Satish, Danielczuk, Sanders, Ichnowski, Lee, Angelova, Vanhoucke, and Goldberg]{huang2021mechanical}
Huang Huang, Marcus Dominguez-Kuhne, Vishal Satish, Michael Danielczuk, Kate Sanders, Jeffrey Ichnowski, Andrew Lee, Anelia Angelova, Vincent Vanhoucke, and Ken Goldberg.
\newblock {Mechanical search on shelves using lateral access x-ray}.
\newblock In \emph{2021 IEEE/RSJ International Conference on Intelligent Robots and Systems (IROS)}, pages 2045--2052. IEEE, 2021.
\newblock \doi{10.1109/IROS51168.2021.9636629.}
\newblock URL \url{https://ieeexplore.ieee.org/document/9636629}.

\bibitem[Jatavallabhula et~al.(2023)Jatavallabhula, Kuwajerwala, Gu, Omama, Chen, Li, Iyer, Saryazdi, Keetha, Tewari, Tenenbaum, de~Melo, Krishna, Paull, Shkurti, and Torralba]{jatavallabhula2023conceptfusion}
Krishna~Murthy Jatavallabhula, Alihusein Kuwajerwala, Qiao Gu, Mohd Omama, Tao Chen, Shuang Li, Ganesh Iyer, Soroush Saryazdi, Nikhil Keetha, Ayush Tewari, Joshua~B Tenenbaum, Celso~Miguel de~Melo, Madhava Krishna, Liam Paull, Florian Shkurti, and Antonio Torralba.
\newblock {ConceptFusion}: Open-set multimodal {3D} mapping, 2023.

\bibitem[Kaelbling et~al.(1998)Kaelbling, Littman, and Cassandra]{KAELBLING199899}
Leslie~Pack Kaelbling, Michael~L Littman, and Anthony~R Cassandra.
\newblock {Planning and acting in partially observable stochastic domains}.
\newblock \emph{Artificial Intelligence}, 101\penalty0 (1):\penalty0 99--134, 1998.
\newblock ISSN 0004-3702.
\newblock \doi{https://doi.org/10.1016/S0004-3702(98)00023-X}.
\newblock URL \url{https://www.sciencedirect.com/science/article/pii/S000437029800023X}.

\bibitem[Kim et~al.(2023)Kim, Kim, Lee, and Park]{kim2023corl}
Seungyeon Kim, Young~Hun Kim, Yonghyeon Lee, and Frank~C Park.
\newblock Leveraging 3d reconstruction for mechanical search on cluttered shelves.
\newblock In \emph{7th Annual Conference on Robot Learning}, 2023.
\newblock URL \url{https://proceedings.mlr.press/v229/kim23a/kim23a.pdf}.

\bibitem[Krainin et~al.(2011)Krainin, Curless, and Fox]{krainin2011autonomous}
Michael Krainin, Brian Curless, and Dieter Fox.
\newblock Autonomous generation of complete 3d object models using next best view manipulation planning.
\newblock In \emph{2011 IEEE international conference on robotics and automation}, pages 5031--5037. IEEE, 2011.

\bibitem[Krause et~al.(2008)Krause, Singh, and Guestrin]{krause2008near}
Andreas Krause, Ajit Singh, and Carlos Guestrin.
\newblock {Near-optimal sensor placements in Gaussian processes: Theory, efficient algorithms and empirical studies.}
\newblock \emph{Journal of Machine Learning Research}, 9\penalty0 (2), 2008.

\bibitem[Lang et~al.(2019)Lang, Vora, Caesar, Zhou, Yang, and Beijbom]{Lang_2019_CVPR}
Alex~H Lang, Sourabh Vora, Holger Caesar, Lubing Zhou, Jiong Yang, and Oscar Beijbom.
\newblock {PointPillars: Fast Encoders for Object Detection From Point Clouds}.
\newblock In \emph{Proceedings of the IEEE/CVF Conference on Computer Vision and Pattern Recognition (CVPR)}, 6 2019.

\bibitem[Li et~al.(2022)Li, Weinberger, Belongie, Koltun, and Ranftl]{li2022languagedriven}
Boyi Li, Kilian~Q Weinberger, Serge Belongie, Vladlen Koltun, and René Ranftl.
\newblock Language-driven semantic segmentation.
\newblock \emph{arXiv preprint arXiv:2201.03546}, 2022.

\bibitem[Li et~al.(2016)Li, Hsu, and Lee]{li2016iros}
Jue~Kun Li, David Hsu, and Wee~Sun Lee.
\newblock Act to see and see to act: Pomdp planning for objects search in clutter.
\newblock In \emph{Proc.~of the IEEE/RSJ Intl.~Conf.~on Intelligent Robots and Systems (IROS)}. IEEE, 2016.
\newblock \doi{10.1109/IROS.2016.7759839}.
\newblock URL \url{https://ieeexplore.ieee.org/document/7759839}.

\bibitem[Liang et~al.(2022)Liang, Xie, Yu, Xia, Lin, Wang, Tang, Wang, and Tang]{NEURIPS2022_43d2b7fb}
Tingting Liang, Hongwei Xie, Kaicheng Yu, Zhongyu Xia, Zhiwei Lin, Yongtao Wang, Tao Tang, Bing Wang, and Zhi Tang.
\newblock {BEVFusion}: A simple and robust lidar-camera fusion framework.
\newblock In S.~Koyejo, S.~Mohamed, A.~Agarwal, D.~Belgrave, K.~Cho, and A.~Oh, editors, \emph{Advances in Neural Information Processing Systems}, volume~35, pages 10421--10434. Curran Associates, Inc., 2022.
\newblock URL \url{https://proceedings.neurips.cc/paper_files/paper/2022/file/43d2b7fbee8431f7cef0d0afed51c691-Paper-Conference.pdf}.

\bibitem[Marques et~al.(2024)Marques, Zhai, Wang, and Hauser]{marques2024overconfidence}
Joao Marcos~Correia Marques, Albert~J Zhai, Shenlong Wang, and Kris Hauser.
\newblock On the overconfidence problem in semantic 3d mapping.
\newblock In \emph{Proc.~of the IEEE Intl.~Conf.~on Robotics \& Automation (ICRA)}. IEEE, 2024.

\bibitem[Monica and Aleotti(2018)]{monica2018contour}
Riccardo Monica and Jacopo Aleotti.
\newblock Contour-based next-best view planning from point cloud segmentation of unknown objects.
\newblock \emph{Autonomous Robots}, 42\penalty0 (2):\penalty0 443--458, 2018.
\newblock URL \url{https://link.springer.com/article/10.1007/s10514-017-9618-0}.

\bibitem[Muguira-Iturralde et~al.(2023)Muguira-Iturralde, Curtis, Du, Kaelbling, and Lozano-Pérez]{visibility-aware_manipulation}
Jose Muguira-Iturralde, Aidan Curtis, Yilun Du, Leslie~Pack Kaelbling, and Tomás Lozano-Pérez.
\newblock Visibility-aware navigation among movable obstacles.
\newblock In \emph{2023 IEEE International Conference on Robotics and Automation (ICRA)}, pages 10083--10089, 2023.
\newblock \doi{10.1109/ICRA48891.2023.10160865}.

\bibitem[Pajarinen et~al.(2023)Pajarinen, Lundell, and Kyrki]{pomdp_compositional_uncertainty}
Joni Pajarinen, Jens Lundell, and Ville Kyrki.
\newblock Pomdp planning under object composition uncertainty: Application to robotic manipulation.
\newblock \emph{IEEE Transactions on Robotics}, 39\penalty0 (1):\penalty0 41--56, 2023.
\newblock \doi{10.1109/TRO.2022.3188168}.

\bibitem[Paszke et~al.(2019)Paszke, Gross, Massa, Lerer, Bradbury, Chanan, Killeen, Lin, Gimelshein, Antiga, Desmaison, Kopf, Yang, DeVito, Raison, Tejani, Chilamkurthy, Steiner, Fang, Bai, and Chintala]{NEURIPS2019_9015}
Adam Paszke, Sam Gross, Francisco Massa, Adam Lerer, James Bradbury, Gregory Chanan, Trevor Killeen, Zeming Lin, Natalia Gimelshein, Luca Antiga, Alban Desmaison, Andreas Kopf, Edward Yang, Zachary DeVito, Martin Raison, Alykhan Tejani, Sasank Chilamkurthy, Benoit Steiner, Lu~Fang, Junjie Bai, and Soumith Chintala.
\newblock {PyTorch: An Imperative Style, High-Performance Deep Learning Library}.
\newblock In H~Wallach, H~Larochelle, A~Beygelzimer, F~d\'{}Alch{\'{e}} Buc, E~Fox, and R~Garnett, editors, \emph{Advances in Neural Information Processing Systems 32}, pages 8024--8035. Curran Associates, Inc., 2019.
\newblock URL \url{http://papers.neurips.cc/paper/9015-pytorch-an-imperative-style-high-performance-deep-learning-library.pdf}.

\bibitem[Pitcher et~al.(2024)Pitcher, F{\"o}rster, and Chung]{pitcher2024iros}
Thomas Pitcher, Julian F{\"o}rster, and Jen~Jen Chung.
\newblock Reinforcement learning for active search and grasp in clutter.
\newblock In \emph{Proc.~of the IEEE/RSJ Intl.~Conf.~on Intelligent Robots and Systems (IROS)}. IEEE, 2024.
\newblock \doi{10.1109/IROS58592.2024.10801366.}
\newblock URL \url{https://ieeexplore.ieee.org/document/10801366}.

\bibitem[Radford et~al.(2021)Radford, Kim, Hallacy, Ramesh, Goh, Agarwal, Sastry, Askell, Mishkin, Clark, Krueger, and Sutskever]{radford2021learning}
Alec Radford, Jong~Wook Kim, Chris Hallacy, Aditya Ramesh, Gabriel Goh, Sandhini Agarwal, Girish Sastry, Amanda Askell, Pamela Mishkin, Jack Clark, Gretchen Krueger, and Ilya Sutskever.
\newblock Learning transferable visual models from natural language supervision, 2021.

\bibitem[Ravi et~al.(2024)Ravi, Gabeur, Hu, Hu, Ryali, Ma, Khedr, R{\"a}dle, Rolland, Gustafson, et~al.]{ravi2024sam}
Nikhila Ravi, Valentin Gabeur, Yuan-Ting Hu, Ronghang Hu, Chaitanya Ryali, Tengyu Ma, Haitham Khedr, Roman R{\"a}dle, Chloe Rolland, Laura Gustafson, et~al.
\newblock {SAM} 2: Segment anything in images and videos.
\newblock \emph{arXiv preprint arXiv:2408.00714}, 2024.

\bibitem[Saxena and Likhachev(2023)]{saxena2023planning}
Dhruv Saxena and Maxim Likhachev.
\newblock Planning for manipulation among movable objects: Deciding which objects go where, in what order, and how.
\newblock In \emph{Proc.~of the Int.~Conf.~on Automated Planning and Scheduling (ICAPS)}, 2023.

\bibitem[Saxena and Likhachev(2024)]{saxena2024improved}
Dhruv~Mauria Saxena and Maxim Likhachev.
\newblock Improved {M4M}: Faster and richer planning for manipulation among movable objects in cluttered 3d workspaces.
\newblock In \emph{Proc.~of the IEEE Intl.~Conf.~on Robotics \& Automation (ICRA)}. IEEE, 2024.
\newblock \doi{10.1109/ICRA57147.2024.10611234.}
\newblock URL \url{https://ieeexplore.ieee.org/document/10611234}.

\bibitem[Sensoy et~al.(2018)Sensoy, Kaplan, and Kandemir]{NEURIPS2018_a981f2b7}
Murat Sensoy, Lance Kaplan, and Melih Kandemir.
\newblock Evidential deep learning to quantify classification uncertainty.
\newblock In S~Bengio, H~Wallach, H~Larochelle, K~Grauman, N~Cesa-Bianchi, and R~Garnett, editors, \emph{Advances in Neural Information Processing Systems}, volume~31. Curran Associates, Inc., 2018.
\newblock URL \url{https://proceedings.neurips.cc/paper_files/paper/2018/file/a981f2b708044d6fb4a71a1463242520-Paper.pdf}.

\bibitem[Sharma et~al.(2023)Sharma, Shivakumar, Huang, Chen, Hoque, Ichter, and Goldberg]{sharma2023openworld}
Satvik Sharma, Kaushik Shivakumar, Huang Huang, Lawrence~Yunliang Chen, Ryan Hoque, Brian Ichter, and Ken Goldberg.
\newblock Open-world semantic mechanical search with large vision and language models.
\newblock In \emph{7th Annual Conference on Robot Learning}, 2023.
\newblock URL \url{https://openreview.net/forum?id=vsEWu6mMUhB}.

\bibitem[Stilman et~al.(2007)Stilman, Schamburek, Kuffner, and Asfour]{stilman2007manipulation}
Mike Stilman, Jan-Ullrich Schamburek, James Kuffner, and Tamim Asfour.
\newblock Manipulation planning among movable obstacles.
\newblock In \emph{Proc.~of the IEEE Intl.~Conf.~on Robotics \& Automation (ICRA)}. IEEE, 2007.

\bibitem[Thrun et~al.(2005)Thrun, Burgard, and Fox]{thrun2005probabilistic}
Sebastian Thrun, Wolfram Burgard, and Dieter Fox.
\newblock {Probabilistic robotics. 2005}.
\newblock \emph{Massachusetts Institute of Technology, USA}, 2005.

\bibitem[Ulmer et~al.(2023)Ulmer, Hardmeier, and Frellsen]{ulmer2023prior}
Dennis Ulmer, Christian Hardmeier, and Jes Frellsen.
\newblock Prior and posterior networks: A survey on evidential deep learning methods for uncertainty estimation.
\newblock \emph{Proceedings of Machine Learning Research}, 2023.

\bibitem[Valevski et~al.(2024)Valevski, Leviathan, Arar, and Fruchter]{valevski2024diffusionmodelsrealtimegame}
Dani Valevski, Yaniv Leviathan, Moab Arar, and Shlomi Fruchter.
\newblock Diffusion models are real-time game engines, 2024.
\newblock URL \url{https://arxiv.org/abs/2408.14837}.

\bibitem[Xiao et~al.(2019)Xiao, Katt, ten Pas, Chen, and Amato]{xiao2019online}
Yuchen Xiao, Sammie Katt, Andreas ten Pas, Shengjian Chen, and Christopher Amato.
\newblock Online planning for target object search in clutter under partial observability.
\newblock In \emph{Proc.~of the IEEE Intl.~Conf.~on Robotics \& Automation (ICRA)}. IEEE, 2019.
\newblock \doi{10.1109/ICRA.2019.8793494.}
\newblock URL \url{https://ieeexplore.ieee.org/document/8793494}.

\bibitem[Yang et~al.(2024)Yang, Du, Ghasemipour, Tompson, Kaelbling, Schuurmans, and Abbeel]{yang2024learning}
Sherry Yang, Yilun Du, Seyed Kamyar~Seyed Ghasemipour, Jonathan Tompson, Leslie~Pack Kaelbling, Dale Schuurmans, and Pieter Abbeel.
\newblock Learning interactive real-world simulators.
\newblock In \emph{The Twelfth International Conference on Learning Representations}, 2024.
\newblock URL \url{https://openreview.net/forum?id=sFyTZEqmUY}.

\bibitem[Zeng et~al.(2020)Zeng, Wen, Zhao, and Liu]{zeng2020view}
Rui Zeng, Yuhui Wen, Wang Zhao, and Yong-Jin Liu.
\newblock View planning in robot active vision: A survey of systems, algorithms, and applications.
\newblock \emph{Computational Visual Media}, 2020.
\newblock URL \url{https://link.springer.com/article/10.1007/s41095-020-0179-3}.

\bibitem[Zhai and Wang(2023)]{Zhai_2023_ICCV}
Albert~J Zhai and Shenlong Wang.
\newblock {PEANUT}: Predicting and navigating to unseen targets.
\newblock In \emph{Proceedings of the IEEE/CVF International Conference on Computer Vision (ICCV)}, pages 10926--10935, October 2023.
\newblock \doi{10.1109/ICCV51070.2023.01003.}
\newblock URL \url{https://ieeexplore.ieee.org/document/10378364}.

\end{thebibliography}
\newpage\clearpage
\appendix
\section{Appendix}




\subsection{Dataset Generation Details}
\label{sec:dataset_generation}
\begin{figure}[!ht]
\centering
\includegraphics[width=.65\linewidth]{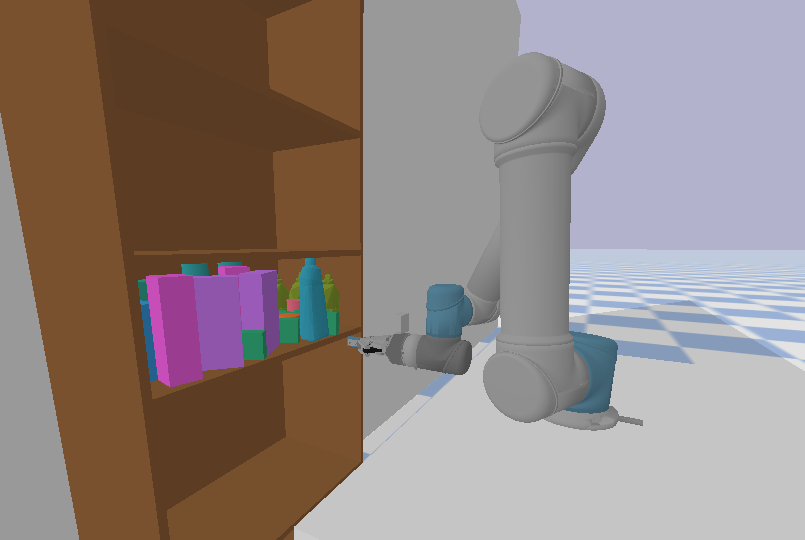}
\caption{Simulation environment configuration example of different YCB objects in a confined shelf scenario. The UR5 is equipped with an Robotiq parallel-jaw gripper}
\label{fig:simulation_setup_figure}
\end{figure}

The simulation environment used for data generation and simulation testing is PyBullet \cite{coumans2021} (\autoref{fig:simulation_setup_figure}).
The object arrangements for the CNABU training datasets were created according to the following procedure: First, we sample the desired occupancy fraction of the shelf floor plan - which we set to be between 30 and 45\%. We define two parameters for each object class: its affinity to other classes and its radius of influence. Classes within an object's affinity class and radius of influence have their probability of appearance increased.
Further, we define a regularity parameter, $\rho$ for the generation, from 0 to 1. When this parameter is zero, there is no enforced alignment in the shelf and when it is one, there is a higher probability for objects to be placed directly in line with the centroid of previously placed objects, emulating more orderly arrangements by increasing the probability of sampling areas directly in front or behind already placed objects. 

We begin sampling by placing a fine grid over the floor space of the scene. Then, until the desired shelf occupancy is reached (or a total number of iterations is reached), we sample a point in this grid that is not yet occupied by another object - taking into consideration the altered probabilities of occupancy by $\rho$. Through the use of Minkowski differences between the object shapes and the free space, we determine the placeable area for the centroid of each object class within the current arrangement and, for each object that is placeable in the selected point, we compute its sampling probability conditioned on the affinities of the objects whose area of influence contains the sampled point. We then sample an object class according to that distribution and randomly sample an angle for the object, between $0^\circ$ and $180^\circ$ and place the object in that orientation. To stimulate the presence of occlusions, we set the base probability of larger objects to be slightly higher - and set large objects to have affinity for smaller objects. For dataset generation, we leave $\rho$ at 0.
The ground truth map for each scenario is collected by removing the top shelf, taking a series of dense top-down images of the scene and mapping them with traditional metric-semantic grid mapping \cite{marques2024overconfidence}. For the pushing dataset, the scenario sampling is identical, except we also sample a random push following \autoref{sec:push_sampling}. 

\begin{figure}[htbp!]
\centering
\subfloat[High Occupancy. sample - frontal]{\includegraphics[width=.5\linewidth, trim={0cm 0cm 0cm 0cm},clip]{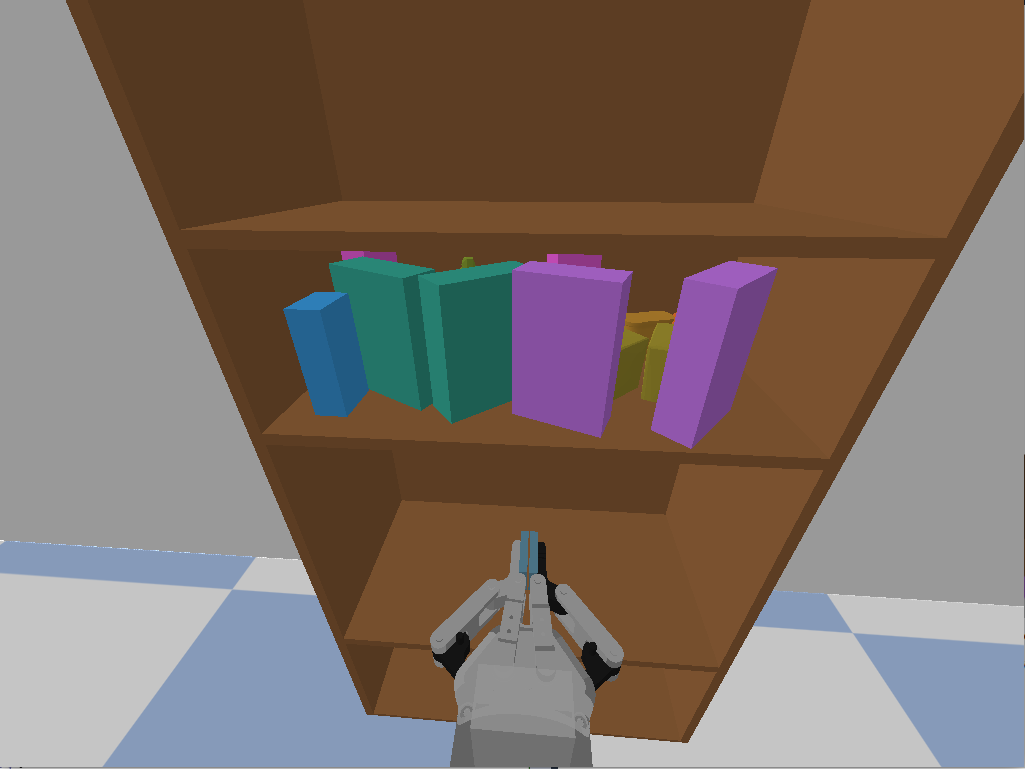}}
\hfill
\subfloat[High Occupancy sample - top]{\includegraphics[width=.5\linewidth,trim={0cm 0cm 0cm 0cm},clip]{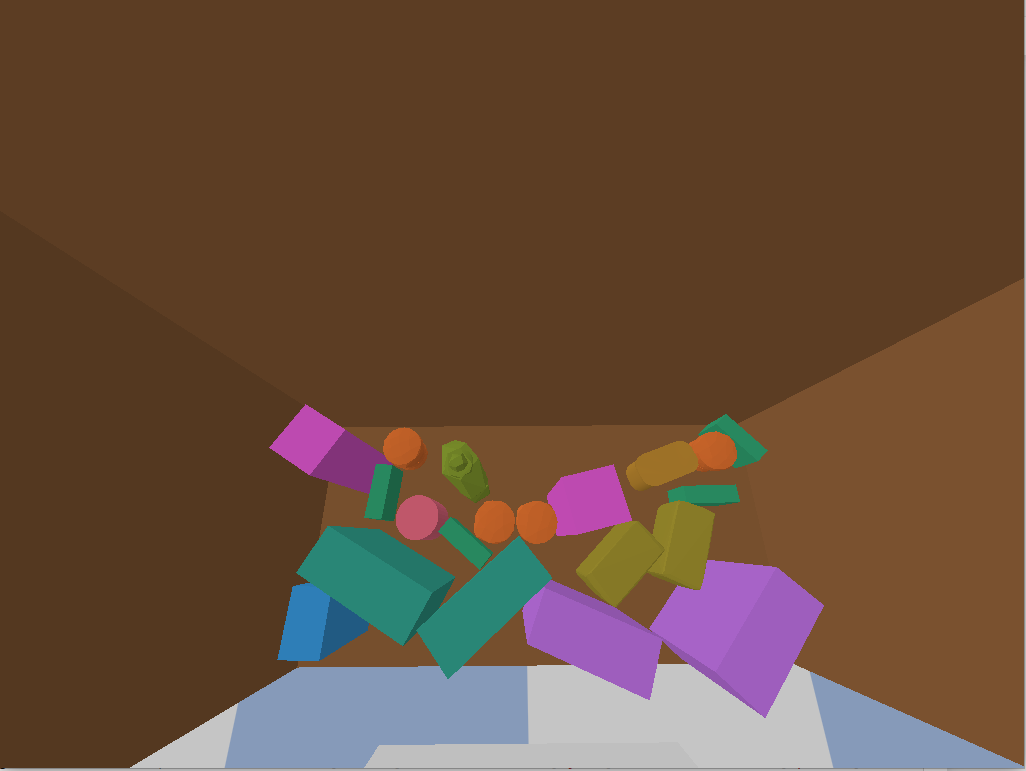}} 
\caption{Frontal and (privileged) top-down views of a scene from the highly occluded set}
\label{fig:simulated_example_hard}

\end{figure}

\begin{figure}[ht!]
\subfloat[Low Occupancy sample - frontal]{\includegraphics[width=.5\linewidth, trim={0cm 0cm 0cm 0cm},clip]{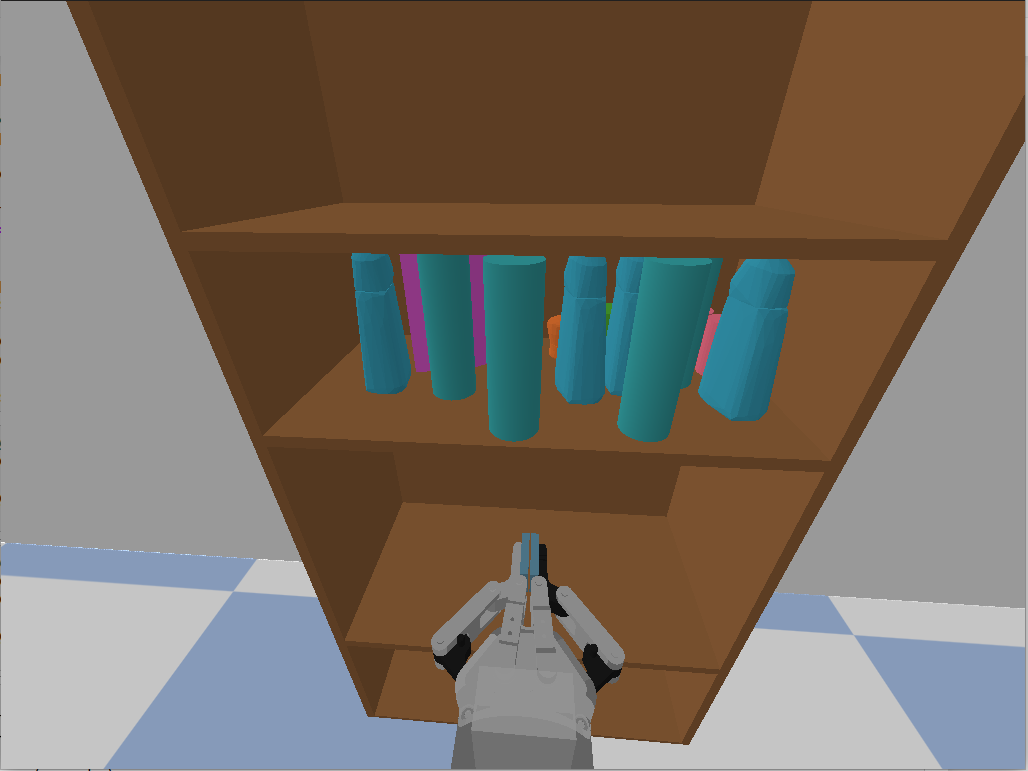}}
\hfill
\subfloat[Low Occupancy sample - top]{\includegraphics[width=.5\linewidth,trim={0cm 0cm 0cm 0cm},clip]{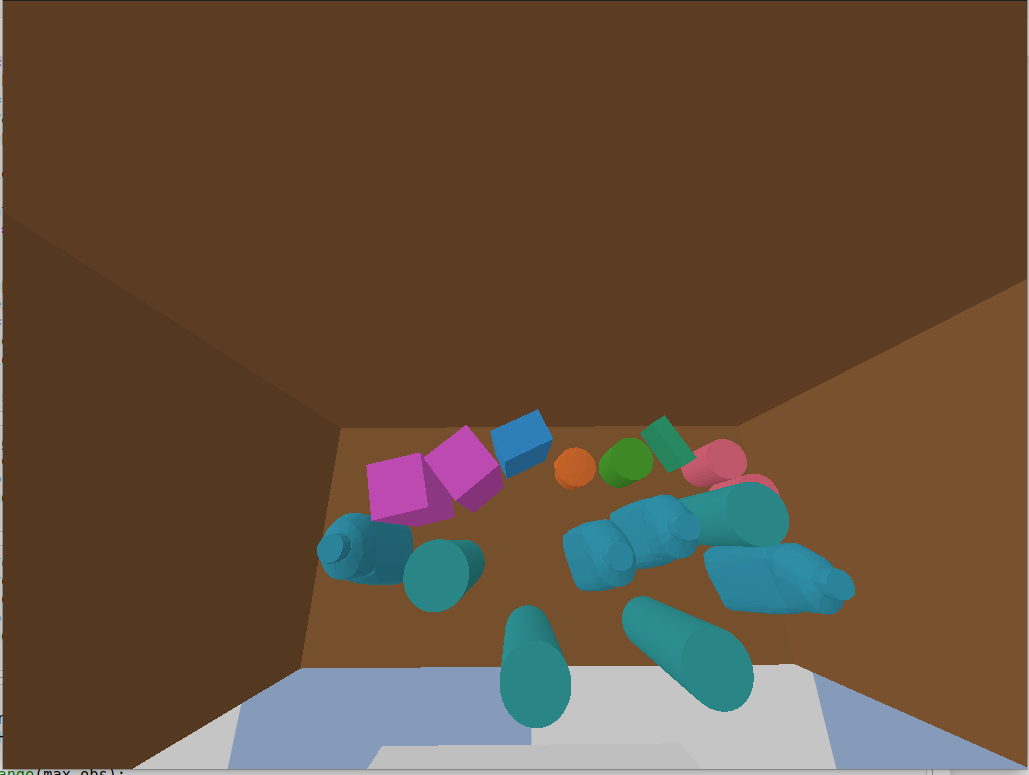}} 
\caption{Frontal and (privileged) top-down views of a scene from the Slightly Occluded set (right). }
\label{fig:simulated_example_easy}

\end{figure}

\begin{table*}[htbp!]
  \centering
  \caption{Summary of features of all considered baselines}
  \begin{tabular}{l|lllll}
    \textbf{Baseline Name} & \textbf{Use $\sigma_o$} & \textbf{VPP} & \textbf{Use $\sigma_m$} & \textbf{Push} & \textbf{Action Selection} \\ \hline
    \textbf{Ours} & yes & yes & yes & yes & \autoref{sec:solving_the_pomdp} \\ 
    \textbf{Ours w/o Pushing} & yes & yes & no & no & no \\ 
    \textbf{Random + Observation CNABU} & yes & no & no & no & no \\ 
    \textbf{Random} & no & no & no & no & no \\ 
    \textbf{Random Push Every $5^{th}$} & yes & yes & yes & yes & Random every 5 steps \\ 
    \textbf{Random Push Every $5^{th}$, No Push CNABU} & yes & yes & no & yes & Random every 5 Steps \\ 
    \textbf{Random View Random Push Every $5^{th}$} & yes & no & no & yes & Random every 5 steps \\ 
    \textbf{Saturation Pushing} & yes & yes & yes & yes & Random upon VPP saturation \\ 
    \textbf{\citet{dengler2023viewpoint}} & no & yes, RL-Based & no & yes & Push classifier network upon saturation \\ 
  \end{tabular}
  \label{tab:table_of_baselines}
\end{table*}

Figures \ref{fig:simulated_example_hard} and \ref{fig:simulated_example_easy} show example scenes from the High Occupancy and Low Occupancy sets. Note how both scenes have occluded objects which are hard to see from any viewpoint - but the HOS scene has more occluded objects - and a much more challenging manipulation environment to uncover the occlusions. 

\subsection{CNABU Implementation Details}
\label{sec:cnabu_implementation_details}
Each CNABU implements a preprocessing step to encode actions and observations in a representation aligned to the map grid. For the observation CNABU $\sigma_o(\bm{\lambda},a_t,o_t)$, we first project the observation occupancies and semantics into an aligned map space using the chosen viewpoint (Free Cell Obs., Occ. Cell Obs., and Semantic Obs. in \autoref{fig:net_architecture}). We then learn $\sigma_o \equiv \sigma_o(\bm{\lambda},\text{Project}(a_t,o_t))$. For the manipulation CNABU, the robot trajectory $a_t$ is projected into an aligned map space that approximates the robot's swept volume. To calculate the robot's swept volume, we approximate the robot's end-effector and last 2 links with simple convex shapes (triangular prism and boxes) and subdivide the robot's path, marking all voxels within those primitives as being part of the swept volume, which is then encoded in a binary 3D voxel grid (swept volume in \autoref{fig:net_architecture}). Additionally, the trajectory start and end points are encoded in 2D binary masks (Start Point and End Point maps in \autoref{fig:net_architecture}).  Ultimately, we learn $\sigma_m \equiv \sigma_m(\bm{\lambda},\text{RobotOccupancy}(a_t(t_s)),\text{RobotOccupancy}(a_t(t_e)))$.

We use network architectures similar to \mbox{\citet{georgakis2022learning}}, with the exception that the output heads are set to be posterior networks. Our network structure can be seen in \autoref{fig:net_architecture} 
The occupancy head in both architectures is a 1x1 convolution, while the differences head in the push prediction network is a series of Convolutional + ReLU + BatchNorm layers followed by a 1x1 convolution. The semantic head is similar to the differences head.  The 2D projection block merely takes a horizontal slice of the occupancy tensor at a fixed height, in our case, 3cm above the shelf.
Their losses and training are described in \autoref{sec:NetworkTraining}. 
We used BEVFusion's approach of feeding the voxel heights as additional channels to 2D Res-UNets for processing the voxel grids as inputs~\cite{NEURIPS2022_43d2b7fb,Lang_2019_CVPR}.

The networks are trained using backpropagation in PyTorch \cite{NEURIPS2019_9015}, with grid search-optimized learning rates and ADAM optimizer, as well as early stopping based on the validation loss. 
The dataset for training $\sigma_o$ consists of 30.000 randomly sampled scenes, while the dataset for training $\sigma_m$ consists of 11.700 pushes. Both datasets were split into train, validation and test splits at a ratio of 0.8:0.1:0.1. Dataset generation details are discussed in \autoref{sec:dataset_generation} of the appendix.
For added robustness in real-world scenarios, we augment the simulation data with salt-and-pepper noise, random rotations and translations and add Gaussian noise to the depth images.

\subsection{Individual CNABU Performance Evaluation}
\label{sec:cnabu_performance_evaluation}

\begin{figure}[htbp!]
\centering
\begin{subfigure}{0.5\textwidth}
\centering
\includegraphics[width=0.8\linewidth, trim={0cm 0cm 0cm 0cm},clip]{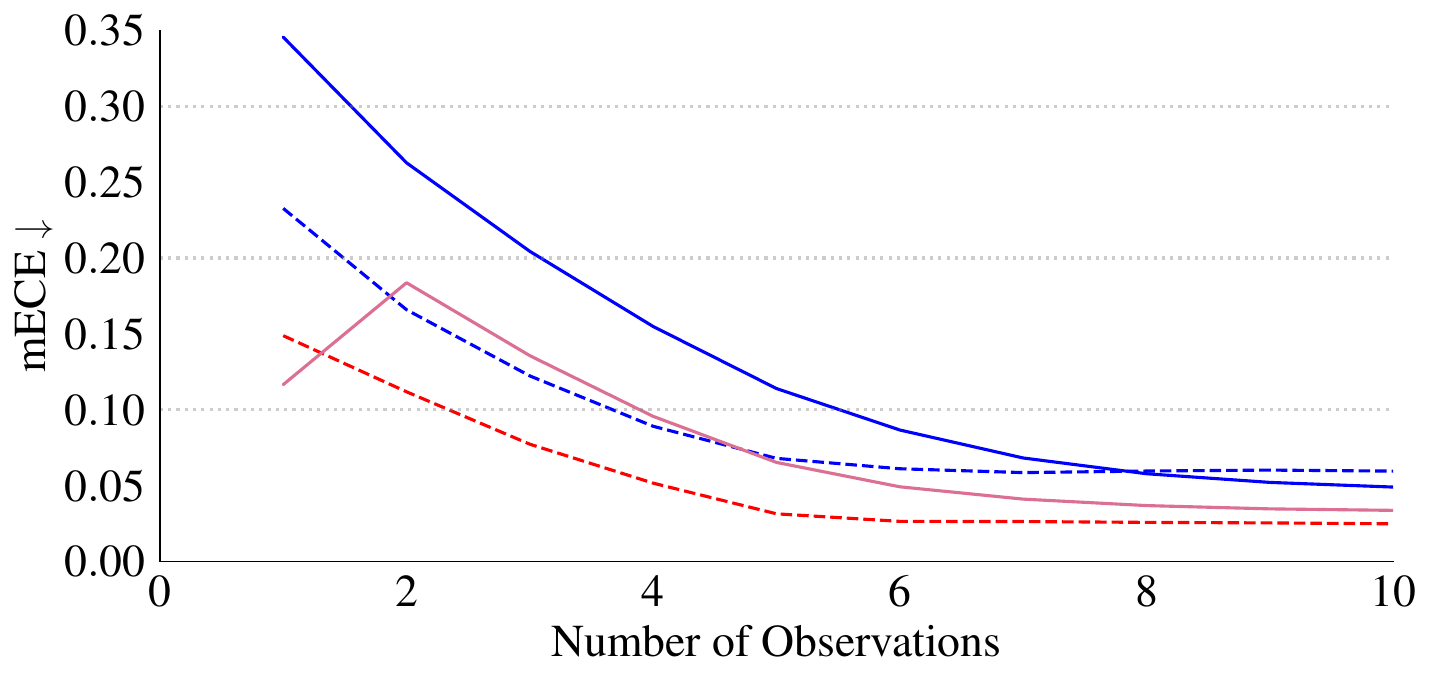}
\caption{mECEs}
\end{subfigure}
\hfill
\begin{subfigure}{0.5\textwidth}
\centering
\includegraphics[width=0.8\linewidth,trim={0cm 0cm 0cm 0cm},clip]{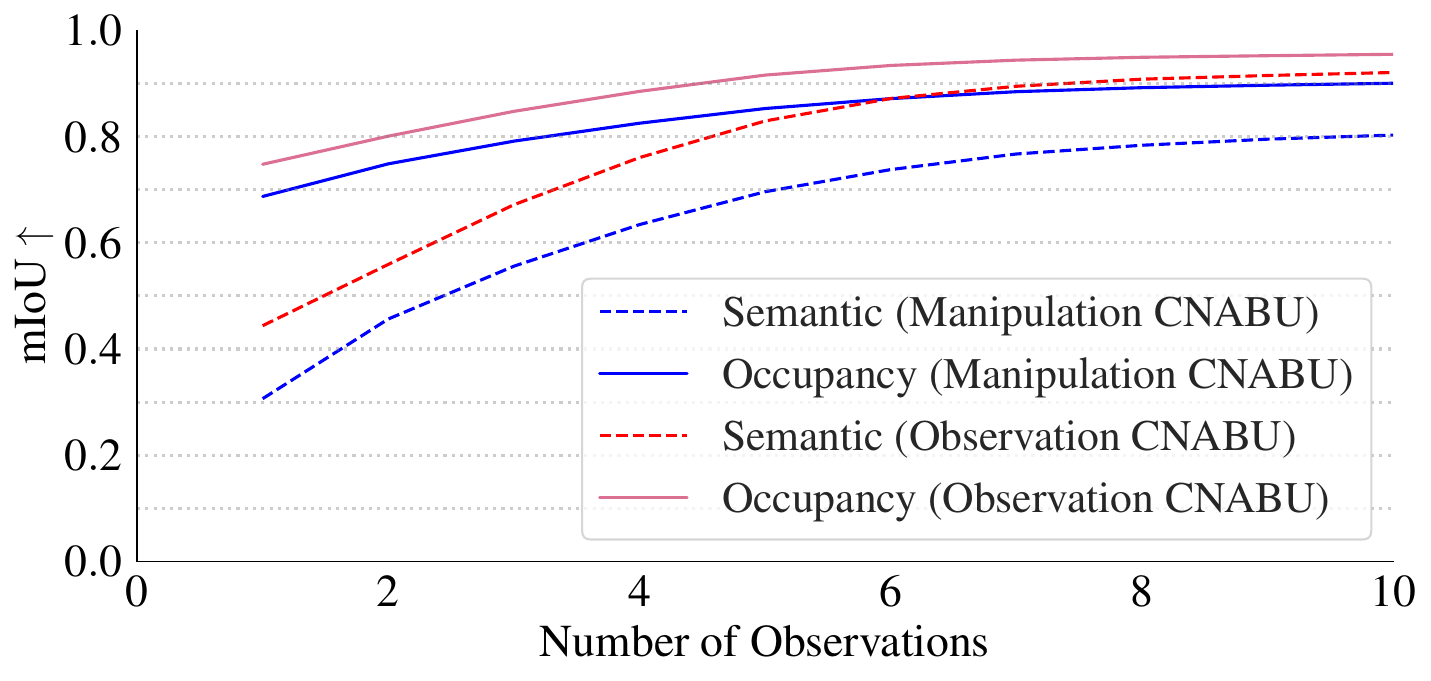}
\caption{mIoUs}
\end{subfigure}
\caption{mECEs and mIoUs for the CNABUs on their respective test sets.}
\label{fig:calibration}
\end{figure}

To evaluate the performance of the trained CNABUs, we use the unseen test set of the dataset used for their training. To evaluate the observation CNABU $\sigma_o$, we select, for each of the scenes, 10 viewpoints to observe at random and obtain the beliefs at each time step, comparing them against the ground truth map. For evaluating the performance of the manipulation CNABU. $\sigma_m$, we also choose 10 viewpoints at random and obtain the pre-manipulation beliefs at every time step, and then obtain the predicted belief after the manipulation is executed. This evaluates the performance of the manipulation CNABU at different reconstruction steps. For each network, we report their mean Intersection over Union and their mean Expected Calibration Error (mECE) \cite{marques2024overconfidence} for both the semantic and occupancy beliefs in \autoref{fig:calibration}. The mIoU serves as a measure of the correctness of the predicitons, while the mECE measures the confidence calibration of these predictions, \textit{i.e.}, how well the predicted confidences align with actual network performance. Note how after few observations, both networks achieve off-the-shelf reasonable calibration and accuracy, which improves as more views are obtained. It is worth noting that the calibration of the manipulation CNABU with very few observations is low because there is no guarantee that one or two random views would have captured the area being manipulated when predicting the belief after the push.

\subsection{Validating Visual Information Gain}
\label{sec:validating_vig}

\begin{figure}[!ht]
\centering
\includegraphics[width=0.8\linewidth]{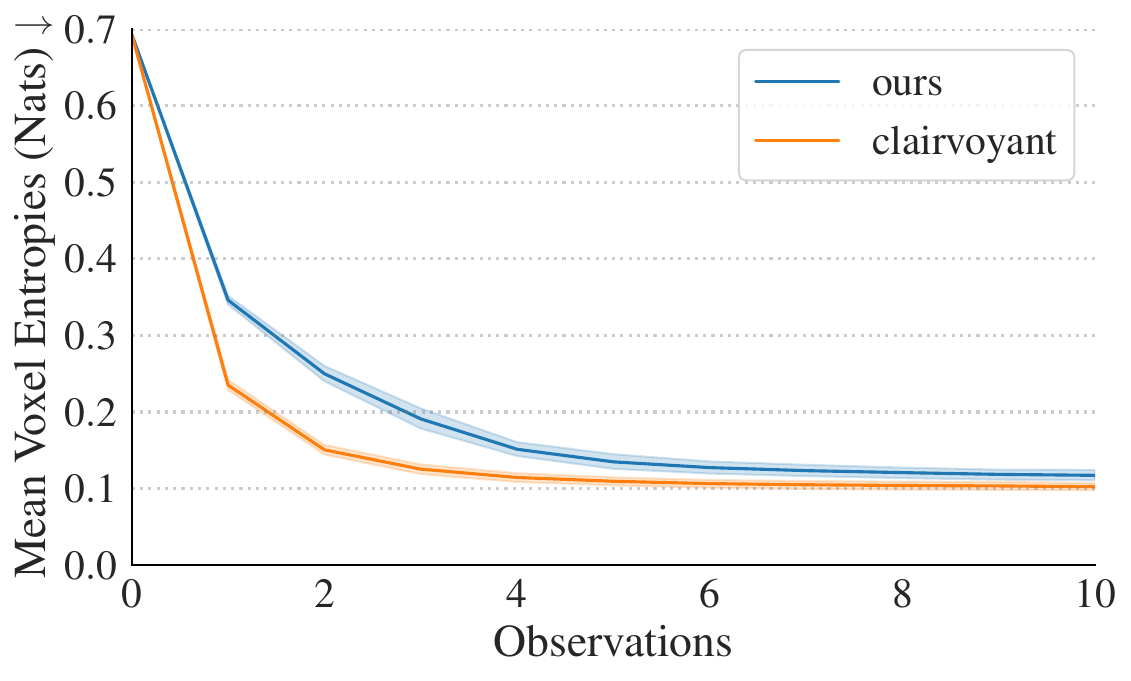}
\caption{Evaluation of the occlusion-aware information gain heuristic against a privileged clairvoyant view selection policy}
\label{fig:clairvoyant_oracle_plot}
\end{figure}

Our proposed planner relies on the assumption that the Volumetric Occlusion-aware Information Gain heuristic, which was developed to estimate information gain in traditional occupancy grid maps, translates well to maps that are no longer fully independent, but predicted via CNABUs. We validate this assumption with the following experiment. Consider the pure Viewpoint Planning task, i.e., we must survey the environment without manipulating it, which is a submodular optimization. Consider now a greedy clairvoyant oracle policy, which, at every time step, has access to all possible observations that could be taken and selects the one that leads to the largest information gain. More formally, this policy is defined by:
\begin{equation}
     v_t^\mathrm{clairvoyant} =  \underset{v\in \mathds{V}}{\arg\max} \left[H(\Phi_{t-1})-H\left(\sigma_o\left(\Phi_{t-1},v\right)\right)\right]
\end{equation}
where $H$ denotes the entropy of the maps.
We compare our agent without pushing to this privileged information agent in the high-occlusion set of scenes and report the resulting mean map occupancy entropies in \autoref{fig:clairvoyant_oracle_plot}. Note how our method which relies on the Occlusion-aware information gain heuristic closely tracks the performance of the privileged clairvoyant oracle in terms of map entropy reduction. While more extensive studies are encouraged, this suggests that this heuristic still performs well, even when the maps are being predicted via CNABUs.

\subsection{Prompts used for Semantic Segmentation}

\label{sec:segmentation_prompts}
To perform semantic segmentation of the SAM2 masks, we use the embeddings of the following text prompts to classify the segments:
\begin{itemize}
  \item tomato can -``tomato or kidney beans round tin can"
  \item chewing gum -``small round chewing gum box" 
  \item spam: "potted meat tin can, spam"
  \item ohio cookies - ``ohio cookie box in purple cardboard carton cookies",
  \item mustard bottle - ``yellow frenchy's mustard bottle",
  \item coffee can - ``maxwell house coffee can with blue wrapper",
  \item gelatin box - ``light pink gelatin box",
  \item cheez-it box - ``cheezeit cracker box in dark red color",
  \item pringles can - ``pringles chips tube cyllinder red or green, in red color or green color bottles with transparent lid",
  \item glass cleaner - ``glass cleaner spray plastic bottle", 
  \item baking powder - ``koop mans baking powder box",
  \item pasta box - ``Big blue carton of pasta collezione" 
  \item Cornflakes - ``kölln schoko cronflakes in yellow brown cardboard box cereal"  
  \item milk carton - ``milk box tetrapak blue label voll milch",
  \item shelf - ``wooden shelf board","cream colored wooden shelf of light cream color"
  \item black - ``just black"
\end{itemize}

\begin{figure}[htbp!]
\centering
\includegraphics[width=.98\linewidth,trim={0.0cm 0.0cm 3cm 0cm},clip]{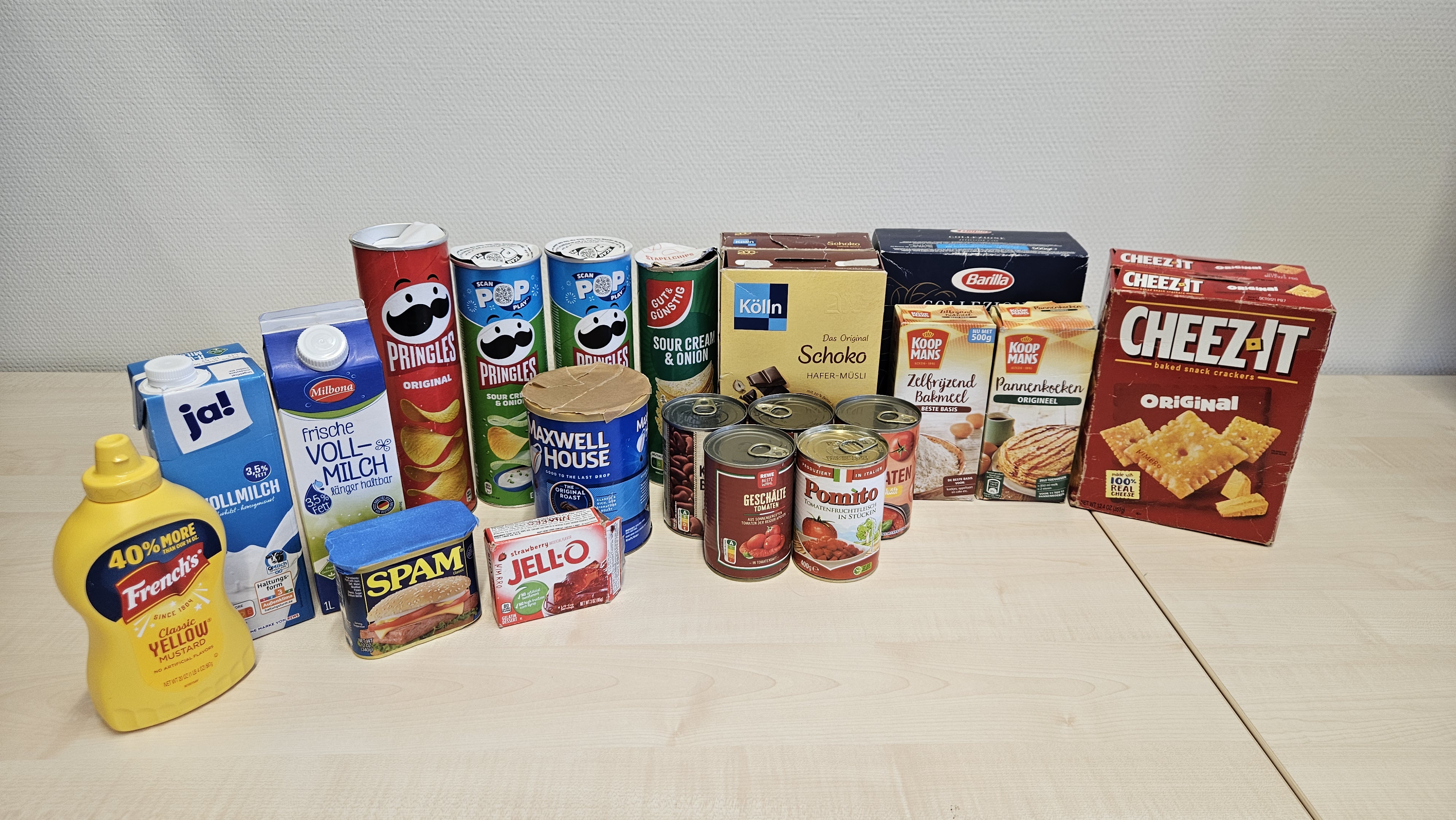}
\caption{The set of objects used during the real world experiments}
\label{fig:real_object_set}
\end{figure}

In this case, both ``shelf" and ``black" were used as synonymous of the background class, capturing different failure cases of SAM2 segmentation. 

\subsection{Summary of baseline features}
\label{sec:baselines_table}
We summarize the considered baselines in \autoref{tab:table_of_baselines} to facilitate comparing their features.

\subsection{Physical Experiments Object Set}

\autoref{fig:real_object_set} shows the objects which were used to create the test scenes during our real world experiments. Note how there are multiple shapes for objects of the same class, like pringles cans, milk cartons and cans. They also differ from the geometries used during training.

\end{document}